\documentclass{article}

\makeatletter
\def\input@path{{styles/}}
\makeatother
\usepackage[preprint]{colm2026_conference}
\usepackage{fontspec}

\normalfont
\usepackage{microtype}
\usepackage{graphicx}
\usepackage{trimclip}
\usepackage{xcolor}
\usepackage{booktabs}
\usepackage{array}
\usepackage{colortbl}
\usepackage{float}
\usepackage{tikz}
\usepackage{pgfplots}
\pgfplotsset{compat=1.18}
\usepgfplotslibrary{groupplots}
\usepackage{hyperref}
\usepackage{url}
\usepackage{amsmath} 
\usepackage{multirow}
\usepackage{caption}
\usepackage{colortbl}
\usepackage{float}
\usepackage{pifont}
\usepackage{caption}
\usepackage{tikz}
\usepackage[most]{tcolorbox}
\usepackage{listings}
\usepackage{pgfplots}
\pgfplotsset{compat=1.18}
\usepgfplotslibrary{groupplots}

\usepackage{fontawesome5}
\newcommand{\emailicon}{\textsuperscript{\faEnvelope[regular]}}

\definecolor{promptbg}{RGB}{250,250,250}
\definecolor{promptborder}{RGB}{56,102,255}
\definecolor{prompttitlebg}{RGB}{56,102,255}
\definecolor{prompttitlefg}{RGB}{255,255,255}
\definecolor{promptleftbar}{RGB}{80,140,120}

\lstdefinestyle{promptstyle}{
    basicstyle=\ttfamily\footnotesize,
    breaklines=true,
    backgroundcolor=\color{promptbg},
    frame=none, numbers=none,
    columns=fullflexible,
    keepspaces=true,
    showspaces=false,
    showstringspaces=false,
    tabsize=2,
    escapeinside=||,
    lineskip=-0.5pt,
}

\newtcolorbox{promptbox}[1]{
    enhanced,
    arc=2pt, boxrule=0.6pt,
    colframe=promptleftbar!50!promptborder,
    colback=promptbg,
    coltitle=prompttitlefg,
    fonttitle=\bfseries\scriptsize,
    title={\strut #1},
    titlerule=0pt,
    toptitle=2pt, bottomtitle=1pt,
    top=3pt, bottom=3pt, left=6pt, right=5pt,
    boxsep=0pt,
    colbacktitle=prompttitlebg,
    before skip=4pt,
    after skip=4pt,
}

\definecolor{abyss}{HTML}{121D36}
\definecolor{polarnight}{HTML}{1A2947}
\definecolor{nebula}{HTML}{2B3F66}
\definecolor{steeltrail}{HTML}{6D87BD}
\definecolor{skytrail}{HTML}{8FA8D8}
\definecolor{starlight}{HTML}{DFE7F5}
\definecolor{warmstar}{HTML}{E8D9C4}
\definecolor{allsparkwordmark}{HTML}{16233F}
\definecolor{allsparkspark}{HTML}{4A659C}
\definecolor{electricblue}{HTML}{3866FF}
\definecolor{covercream}{HTML}{EEF3FA}
\definecolor{coveraccent}{HTML}{3866FF}
\colorlet{pevekpurple}{skytrail}
\colorlet{bargray}{steeltrail}
\colorlet{barlgray}{starlight}

\newfontfamily\outfit[
  Path=assets/fonts/,
  UprightFont=Outfit-Regular.ttf,
  BoldFont=Outfit-SemiBold.ttf
]{Outfit}

\hypersetup{
  colorlinks=true,
  linkcolor=electricblue,
  citecolor=electricblue,
  urlcolor=coveraccent,
  filecolor=electricblue
}
\setcitestyle{numbers,square,comma,sort&compress}

\newcommand{\reporttitle}{Benchmarking VLMs on Real-Life Questions from Human Communities}
\title{\reporttitle}
\author{AllSpark Team}

\begin{document}

\fancyhead{}
\renewcommand{\headrulewidth}{0pt}
\color{abyss}
\thispagestyle{empty}

\vspace*{-0.44in}
\begin{tcolorbox}[
  width=\linewidth,
  colback=covercream,
  colframe=covercream,
  boxrule=0pt,
  arc=14pt,
  outer arc=14pt,
  boxsep=0pt,
  left=20pt,
  right=20pt,
  top=13pt,
  bottom=11pt
]
  {\outfit\fontsize{21.5}{25.5}\selectfont\bfseries\centering
    \textcolor{coveraccent}{NoteVQA}:\hspace{0.25em}\reporttitle\par}
  \vspace{1.45em}
  {\bfseries\centering AllSpark Team\par}

  \vspace{0.75em}
  \begingroup
  \normalfont
  \setlength{\parindent}{0pt}
  \setlength{\parskip}{0pt}
  Vision-language models (VLMs) increasingly power consumer-facing AI search, yet evaluating them on the diversity of everyday visual questions remains challenging. 
Existing benchmarks often target predefined capabilities, such as multi-hop retrieval or long-form synthesis, whereas users ask photo-grounded questions spanning a long tail of everyday scenarios. 
Despite advances in VLMs, users on Xiaohongshu, a mainstream Chinese image-sharing platform, continue to turn to other people for help with everyday visual questions. 
Motivated by this behaviour, we curate NoteVQA from these questions, yielding 252 items across 12 topical categories and 7 user intents. 
Each item includes a concise reference distilled from expert community responses and a human-audited interleaved reference answer that combines textual explanations with supporting visual evidence. 
We evaluate both short-answer correctness and interleaved-answer quality. To support the latter, we introduce AgenticInterleave, a single-agent ReAct framework for retrieval-supported answer generation, together with IVR-12, a 12-dimensional rubric for assessing the content, presentation, and image quality of interleaved references and model outputs.
Across 9 frontier VLMs, the highest short-answer accuracy is 52.8\%, while adding agentic search to Qwen3.5-397B-A17B improves accuracy by only 2.0\%. For interleaved answers, the same model running AgenticInterleave scores 3.52 under IVR-12, compared with 4.65 for the human-audited references, with the largest gap in content quality. These results highlight the challenges that everyday visual questions pose for current VLMs in both answer accuracy and the quality of visually grounded explanations.

  \par
  \endgroup

  \vspace{0.65em}
  \noindent
  \begin{minipage}[b]{0.63\linewidth}
    \outfit\fontsize{8.4}{10.2}\selectfont
    \textbf{Date:} September 14, 2026\\[-0.1em]
    \textbf{Github:} 
    \url{https://github.com/AllSpark-Research/notevqa}
  \end{minipage}%
  \hfill
  \begin{minipage}[b]{0.33\linewidth}
    \raggedleft
    \raisebox{-0.30em}{\includegraphics[height=16pt]{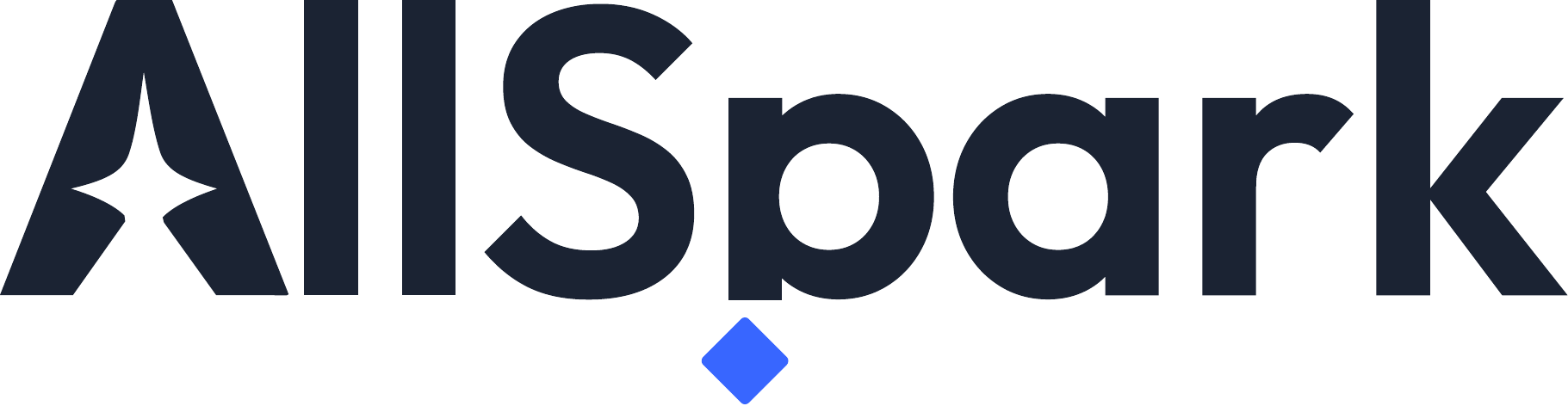}}%
  \end{minipage}
\end{tcolorbox}

\vspace{0.1em}
\begin{figure}[H]
\centering
\includegraphics[width=\linewidth]{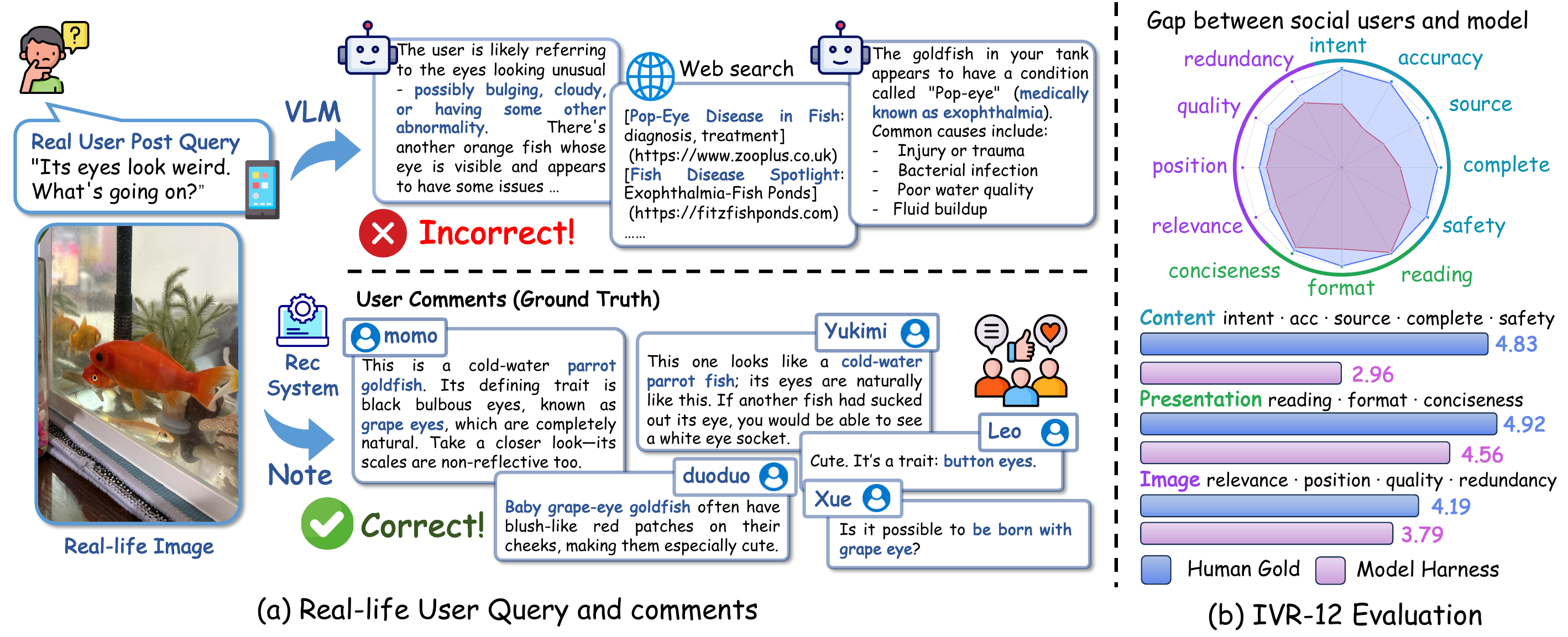}
\caption{Overview of NoteVQA.
(a) A real-user visual query that remains challenging for a VLM with web retrieval but is correctly answered by community experts. (b) IVR-12 comparison of the human-audited gold reference and AgenticInterleave output across content, presentation, and image quality.
}
\label{fig:teaser}
\end{figure}
\clearpage

\section{Introduction}
\label{sec:intro}

Vision-language models (VLMs) are rapidly becoming a new interface for information seeking, allowing users to ask questions directly about the visual world around them. Yet many existing benchmarks follow a capability-first paradigm: researchers define a target capability, such as compositional visual reasoning, knowledge-based question answering, or multimodal retrieval, and then construct questions to evaluate it~\cite{johnson2017clevr,hudson2019gqa,marino2019ok,chang2022webqa,he2026vistahop}. While methodologically sound, this approach can produce queries that differ substantially from those users submit in practice. AI search systems such as Perplexity~\cite{perplexity} and those embedded in user-generated content (UGC) platforms must address diverse everyday information needs. In image-based communities, users often pair a photo with a brief, naturally phrased question spanning the long tail of everyday scenarios. Strong performance on researcher-authored benchmarks may therefore not reliably translate into usefulness for real users.

We introduce NoteVQA, a benchmark built from visual questions posted by users on Xiaohongshu~\footnote{\url{https://www.xiaohongshu.com/}}, a mainstream Chinese platform for user-generated image posts. In the posts collected for NoteVQA, users share photos, ask questions, and seek answers from other community members. These questions provide a concrete setting for examining how well current VLMs address real-world visual information needs. Prior work identifies fine-grained entity recognition and knowledge-intensive visual questions as challenges for VLMs~\cite{chen2023infoseek}, motivating our focus on visual identification and local or niche knowledge. We also cover subjective judgment and authenticity checking to capture a broader range of everyday needs. Studies of social question answering highlight trust, personalised advice, and subjective recommendations as reasons for asking other people~\cite{morris2010social}, underscoring the importance of answer usefulness alongside factual accuracy. Community question-answering research further associates topic-specific expertise with higher answer ratings in factual categories~\cite{adamic2008knowledge}. We therefore draw on community replies as a source of candidate answers and refine them through expert review, as illustrated in Figure~\ref{fig:teaser}(a).

Beyond the source of the questions, the answer format also matters. Benchmarks such as SimpleVQA~\cite{cheng2025simplevqa}, FVQA~\cite{wang2017fvqa}, MMSearch~\cite{jiang2024mmsearch}, and BrowseComp-VL (BC-VL)~\cite{geng2026webwatcher} use short reference answers, with median lengths of only a few words. Short-answer accuracy measures whether a model reaches the correct answer, but does not directly assess the quality of its explanation or supporting visual evidence. Community questions often call for more than identification: users also seek explanations, actionable guidance, and help distinguishing visually similar objects. To evaluate how well models meet these needs, NoteVQA pairs short references distilled from community responses with human-audited interleaved reference answers that combine textual explanations with supporting images. These two reference formats enable complementary evaluations of answer correctness and visually grounded explanation quality on the same questions.

Evaluating interleaved answers requires both a generation framework that can retrieve and incorporate visual evidence and criteria that assess how effectively the resulting text and images answer the question. We provide AgenticInterleave, a single-agent ReAct~\cite{yao2022react} framework that generates retrieval-supported interleaved answers with traceable image tags. We also introduce IVR-12, a 12-dimensional rubric covering content, presentation, and image quality. Applying the same criteria to human-audited references and model outputs provides a common scale for examining where generated answers succeed and fall short.

Our main contributions are as follows.

\begin{itemize}

\item \textbf{A community-sourced visual QA benchmark with two reference formats.} We construct NoteVQA, comprising 252 real-world questions across 12 topical categories and 7 user intents. Each question is paired with an expert-distilled short reference and a human-audited interleaved reference, supporting evaluation of both answer correctness and visually grounded explanation quality.

\item \textbf{A framework and rubric for interleaved answer evaluation.} We provide AgenticInterleave for generating retrieval-supported interleaved answers with traceable image tags, together with IVR-12 for assessing content, presentation, and image quality under the same criteria for references and model outputs.

\item \textbf{An empirical analysis of accuracy, retrieval, and image use.} We evaluate 9 frontier VLMs, with the strongest achieving only 52.8\% short-answer accuracy. Agentic retrieval improves Qwen3.5-397B-A17B by just 2.0 percentage points, while its interleaved answers show a substantial content-quality gap relative to the human-audited references. A controlled image-ablation study further quantifies the contribution of embedded visual evidence to answering image-critical probes.

\end{itemize}

\section{Comparison with Existing Multimodal QA Benchmarks}
\label{sec:quant}

We compare NoteVQA with 10 existing open multimodal QA benchmarks. The comparison considers three complementary perspectives: (1) query characteristics and domain coverage, (2) gains from agentic retrieval, and (3) reference-answer formats and evaluation scope. For query characteristics, we report mean query length and estimated reasoning depth. Query length provides a simple indicator of prompt complexity, while the mean number of reasoning hops, estimated by an LLM judge, approximates the depth of retrieval and reasoning required.

\begin{figure}[!ht]
\centering
\color{abyss}
\begin{minipage}[t]{0.60\linewidth}
    \centering
    \captionof{table}{Comparison of open multimodal VQA benchmarks with our proposed NoteVQA across key statistics.}
    \label{tab:motivation}
    \vspace{3pt}
    \setlength{\aboverulesep}{0pt}
    \setlength{\belowrulesep}{0pt}
    \setlength{\tabcolsep}{1.8pt}
    \renewcommand{\arraystretch}{1.02}
    \resizebox{\linewidth}{!}{%
    \begin{tabular}{l r r r l}
    \toprule
    Benchmark &
    Num. &
    Q. len. &
    Hops &
    Domain focus \\
    \midrule
    SimpleVQA~\cite{cheng2025simplevqa}
    & 1013 & 10.2 & 1.52 & Generic perception \\
    FVQA~\cite{wang2017fvqa}
    & 1800 & 11.5 & 2.06 & Factual single-entity QA \\
    InfoSeek~\cite{chen2023infoseek}
    & 2000 & 8.2 & 1.95 & KB entity attributes \\
    MMSearch~\cite{jiang2024mmsearch}
    & 171 & 14.2 & 2.50 & Web-search QA \\
    LiveVQA~\cite{fu2025livevqa}
    & 300 & 52.1 & 2.52 & News timeliness, MCQ \\
    MMSearch+~\cite{tao2026mmsearch}
    & 222 & 12.0 & 2.12 & arXiv multi-image QA \\
    MMBC~\cite{li2025mm}
    & 130 & 56.6 & 3.72 & Web-browsing QA \\
    BC-VL L1~\cite{geng2026webwatcher}
    & 199 & 20.0 & 3.27 & Cross-modal multi-hop \\
    BC-VL L2~\cite{geng2026webwatcher}
    & 200 & 74.0 & 3.88 & Fuzzified long queries \\
    VDR~\cite{zeng2026vision}
    & 500 & 25.9 & 3.05 & Fine-grained perception QA \\
    \midrule
    NoteVQA
    & 252
    & 22.7
    & 2.48
    & Consumer life \\
    \bottomrule
    \end{tabular}}
\end{minipage}\hfill
\begin{minipage}[t]{0.38\linewidth}
    \centering
    \vspace{0pt}
    \includegraphics[width=\linewidth]{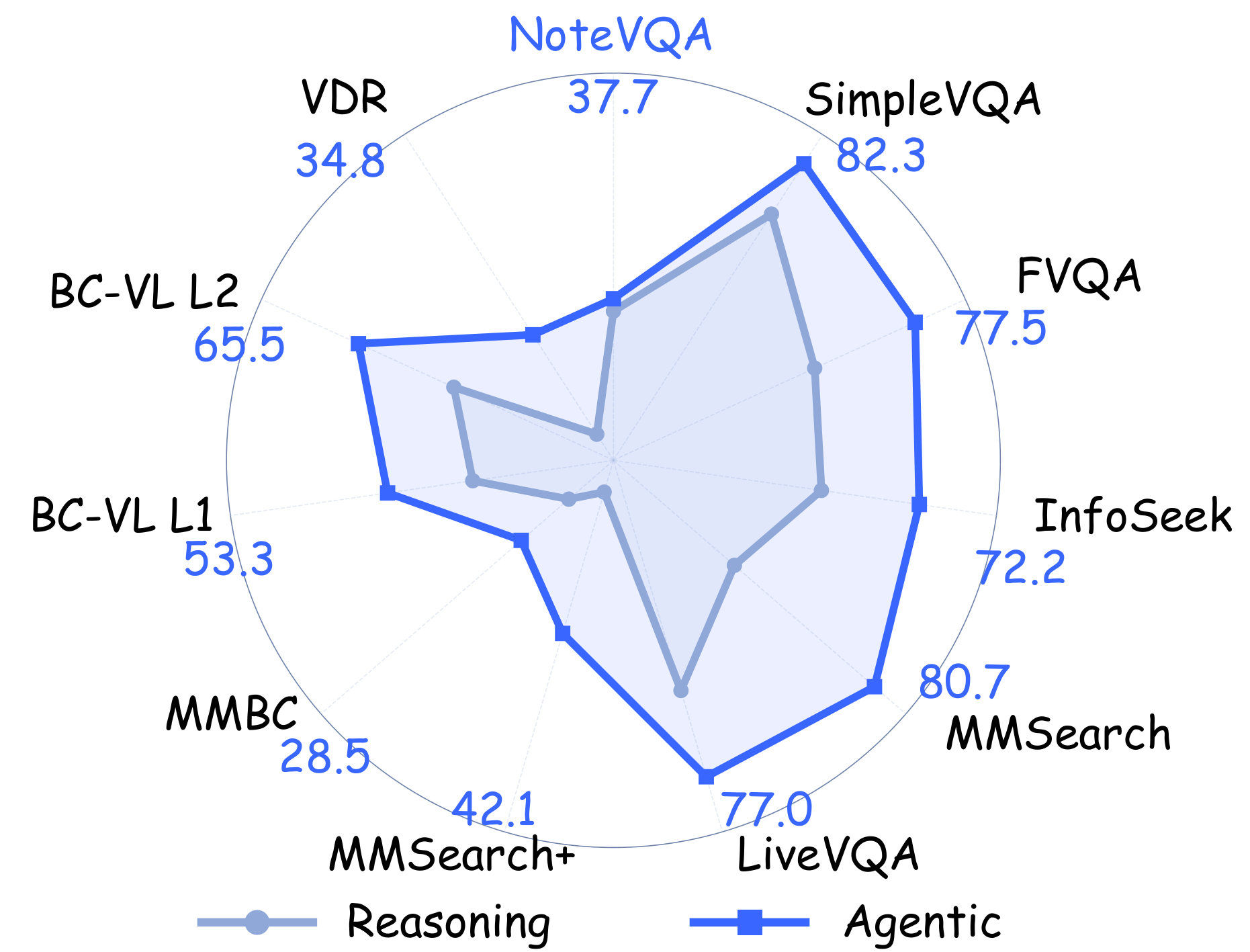}
    \captionof{figure}{Accuracy comparison between reasoning and agentic settings.}
    \label{fig:agentic_gain}
\end{minipage}
\end{figure}

\paragraph{(1) Query Characteristics and Domain Coverage.}
Table~\ref{tab:motivation} compares the query length, estimated reasoning depth, and primary focus of the benchmark settings. BrowseComp-VL L2 and MM-BrowseComp have mean query lengths of 56 to 74 words and estimated reasoning depths of 3.7 to 3.9 hops, reflecting tasks with extended descriptions and explicit reasoning constraints. SimpleVQA, InfoSeek, and MMSearch have shorter queries, averaging 8 to 14 words and 1.5 to 2.5 estimated hops, and primarily assess generic visual perception, entity attributes, or web-search question answering. NoteVQA averages 22.7 words and 2.48 estimated hops, placing it between short factual questions and heavily specified multi-step tasks. Its questions are distinguished by their origin in an online user community and their focus on everyday visual information needs.

\paragraph{(2) Retrieval Gains across Benchmarks.}
Figure~\ref{fig:agentic_gain} compares Qwen3.5-397B-A17B~\cite{team2026qwen3} under two settings on each benchmark. In the single-pass setting, the model answers directly from the query and image without retrieval. In the agentic setting, the same model operates through a single-agent ReAct loop equipped with web search, page visit, and image search. Agentic retrieval improves accuracy by 14 to 43 percentage points across the 10 comparison settings. On NoteVQA, accuracy increases from 35.7\% to 37.7\% (Table~\ref{tab:main}), yielding a substantially smaller gain of 2.0 percentage points. This comparison shows that NoteVQA presents a setting in which the evaluated retrieval system provides limited additional benefit. Section~\ref{sec:exp} analyses the result at the instance level, examining where retrieval helps and where it fails to resolve or introduces errors.

\paragraph{(3) Reference-Answer Formats and Evaluation Scope.}
The comparison benchmarks primarily assess final-answer correctness using concise reference answers or multiple-choice labels. These formats are effective for determining whether a model reaches the expected answer, but do not directly evaluate the quality of a complete explanation supported by visual evidence. Community questions often call for richer responses that combine identification, practical guidance, and visual comparisons that help users distinguish similar objects. NoteVQA therefore provides both short references and human-audited interleaved reference answers. The two formats support complementary evaluations of answer correctness and visually grounded explanation quality on the same set of questions.

Taken together, these comparisons distinguish NoteVQA along three dimensions: it draws everyday visual questions from a real user community, presents limited gains under the evaluated agentic retrieval setting, and supports evaluation of both short and interleaved answers. The following section describes how NoteVQA selects and refines its questions and constructs the two reference formats.

\section{NoteVQA}
\label{sec:notevqa}

Three principles guide NoteVQA. First, questions originate from real Xiaohongshu user posts and are subsequently refined and selected for the benchmark. Second, a vision-necessity gate screens out items judged answerable from the text query alone. Third, reference construction combines community knowledge with human review: short references are distilled from comments by domain experts, while model-generated interleaved drafts are audited by annotators before inclusion.

\subsection{Data Curation Pipeline}
\label{sec:pipeline}

\begin{figure}[t]
\centering
\includegraphics[width=\linewidth]{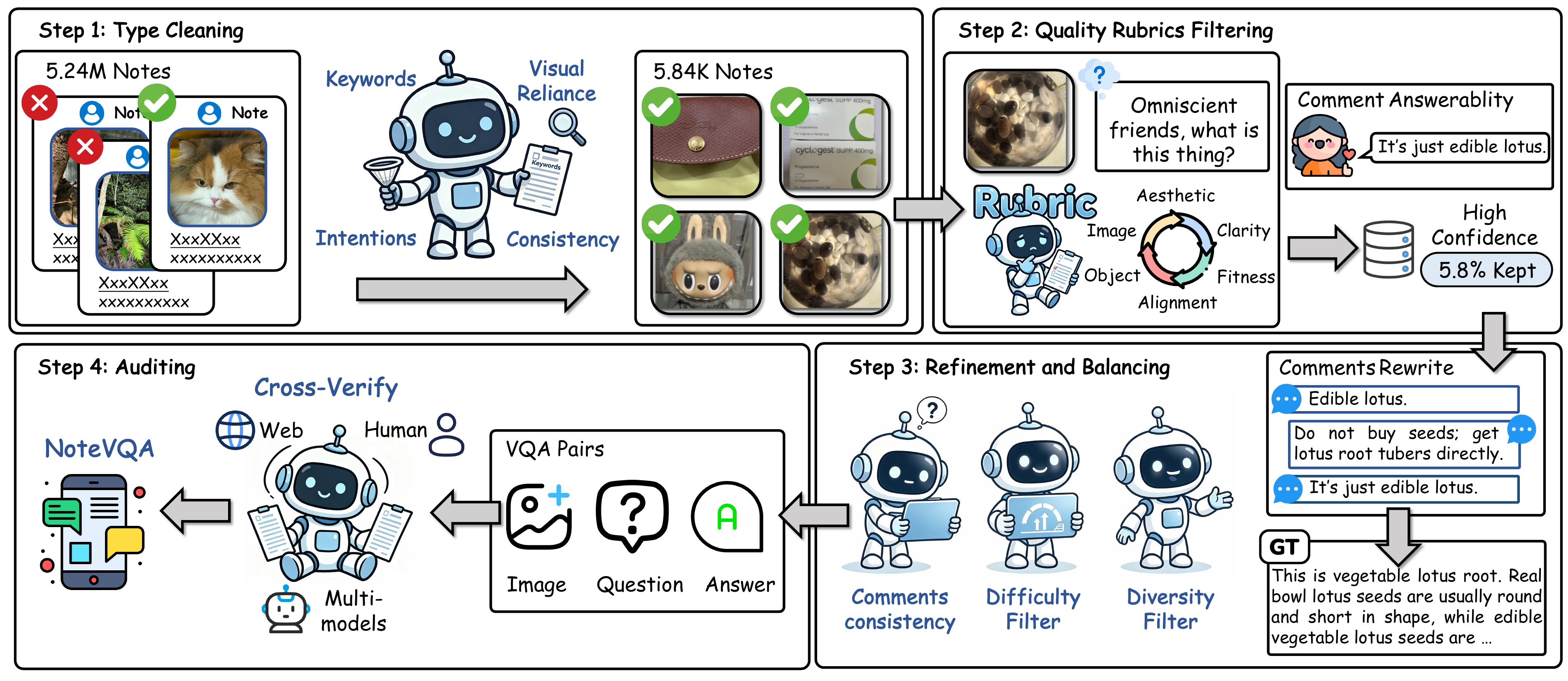}
\caption{Data Curation Pipeline of NoteVQA.
Raw notes flow through Step 1: Type cleaning to yield vision-necessary items, Step 2: Quality filtering to obtain high-confidence samples, Step 3: Refinement and balancing to produce the refined item set, and Step 4: Auditing to produce the final English benchmark.
}
\label{fig:pipeline}
\end{figure}

We implement these principles through the four-step curation pipeline shown in Figure~\ref{fig:pipeline}. 

\paragraph{Step 1: Type Cleaning.} We first apply keyword and length filters to retain image-bearing UGC posts with sufficient textual content, then use a VLM~\cite{team2026qwen3} to identify potential VQA candidates. We further remove items answerable from text alone and verify image--text consistency with a VLM, reducing 5.24M raw notes to 5.84K candidates.

\paragraph{Step 2: Quality Rubric Filtering.} Each candidate is screened by a VLM aesthetic filter and an interleaved-suitability filter requiring the answer to cover at least three substantive aspects. We then apply a five-axis rubric covering question clarity, image quality, image--text alignment, interleaved fit, and answer objectivity, followed by per-tag top-K selection. A final VLM check verifies whether the comments provide a plausible answer, leaving 338 high-confidence samples.

\paragraph{Step 3: Refinement and Balancing.} We standardise query and answer styles and rewrite queries to remove answer-revealing domain terms while preserving the original intent. We then run Qwen3.5-397B-A17B ten times per item and use response consistency as a difficulty probe. Samples are further pruned and balanced across category and intent groups, yielding a final set of 252 items.

\paragraph{Step 4: Auditing.} We translate the refined items into English, cross-check the translations with multiple models, and have human reviewers proofread them for correctness and fluency. Each item is then assigned to its corresponding category and intent cell as a final consistency check.

\begin{figure}[H]
\centering
\includegraphics[width=\linewidth]{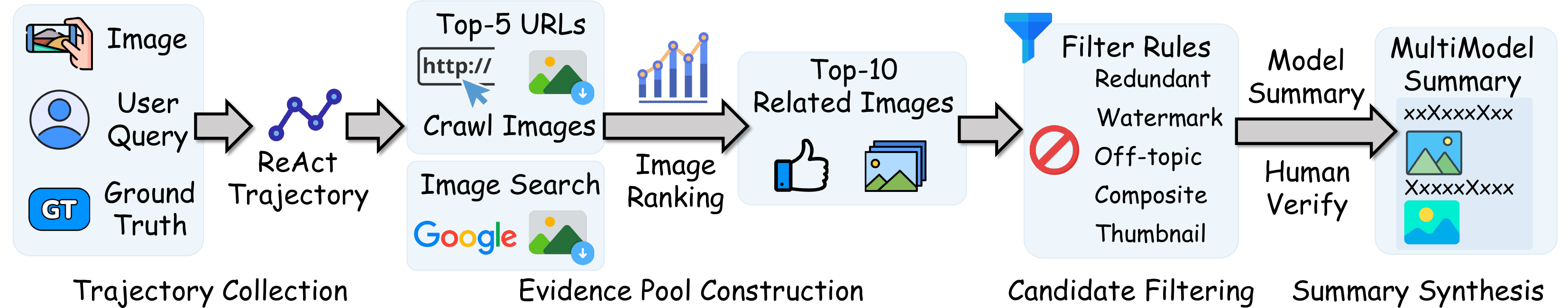}
\caption{Four-stage pipeline for constructing interleaved multimodal summaries.
}
\label{fig:interleaved-pipeline}
\end{figure}

\subsection{Human-Verified Interleaved Reference Construction}
\label{sec:interleaved-gt}

As shown in Figure~\ref{fig:interleaved-pipeline}, we construct a human-verified interleaved reference for each NoteVQA item through a four-stage pipeline: trajectory collection, evidence-pool construction, candidate filtering, and summary synthesis with human audit. The pipeline combines agentic retrieval with multimodal ranking and human verification to select relevant, informative images while retaining their source links.

\paragraph{Stage 1: Trajectory Collection.}
Given the original query image, user query, and short reference answer, a multimodal Qwen3.5-397B-A17B agent runs a ReAct loop using web search, reverse image search, and page visits. The short reference is provided as guidance to encourage evidence collection around the intended answer rather than unconstrained exploration. We retain the resulting tool trajectory and retrieved sources as the evidence record for each item.

\paragraph{Stage 2: Evidence-Pool Construction.}
We construct the candidate evidence pool from two complementary sources. First, an LLM scores all visited URLs along five dimensions---relevance, informativeness, directness, uniqueness, and authoritativeness---and retains the top-5 URLs per item. Images are then crawled from these pages, subject to minimum-size constraints and a cap of 20 images per URL. Second, we preserve images returned by reverse image search during trajectory collection. After merging the two pools, Qwen3.5-397B-A17B scores each image along visual relevance, informativeness, clarity, answer support, and necessity, and retains the top-10 candidates by overall score. When the reference mentions a named entity not represented in the candidate pool, we additionally trigger forward image search to retrieve supplementary visual evidence.

\paragraph{Stage 3: Candidate Filtering.}
Before synthesis, candidate images undergo rule-based and multimodal filtering to remove evidence that is unsuitable for an interleaved reference. We remove redundant images, composites, watermarked images, video thumbnails, and off-topic content, then deduplicate the remaining images. This stage reduces visually noisy or misleading evidence while preserving complementary images that contribute distinct information to the answer.

\paragraph{Stage 4: Summary Synthesis.}
We use Qwen3.5-397B-A17B to combine the original query image, short reference, trajectory evidence, and filtered images into an interleaved multimodal reference, with each embedded image explicitly indexed to its candidate source. The generated reference then undergoes multimodal consistency checks and human review. PhD annotators label each item as \emph{accept}, \emph{minor fix}, or \emph{reject}, correcting content and image placement when necessary. Only references that pass this audit are included in the final benchmark.

\begin{figure}[!ht]
\centering
\includegraphics[width=\linewidth]{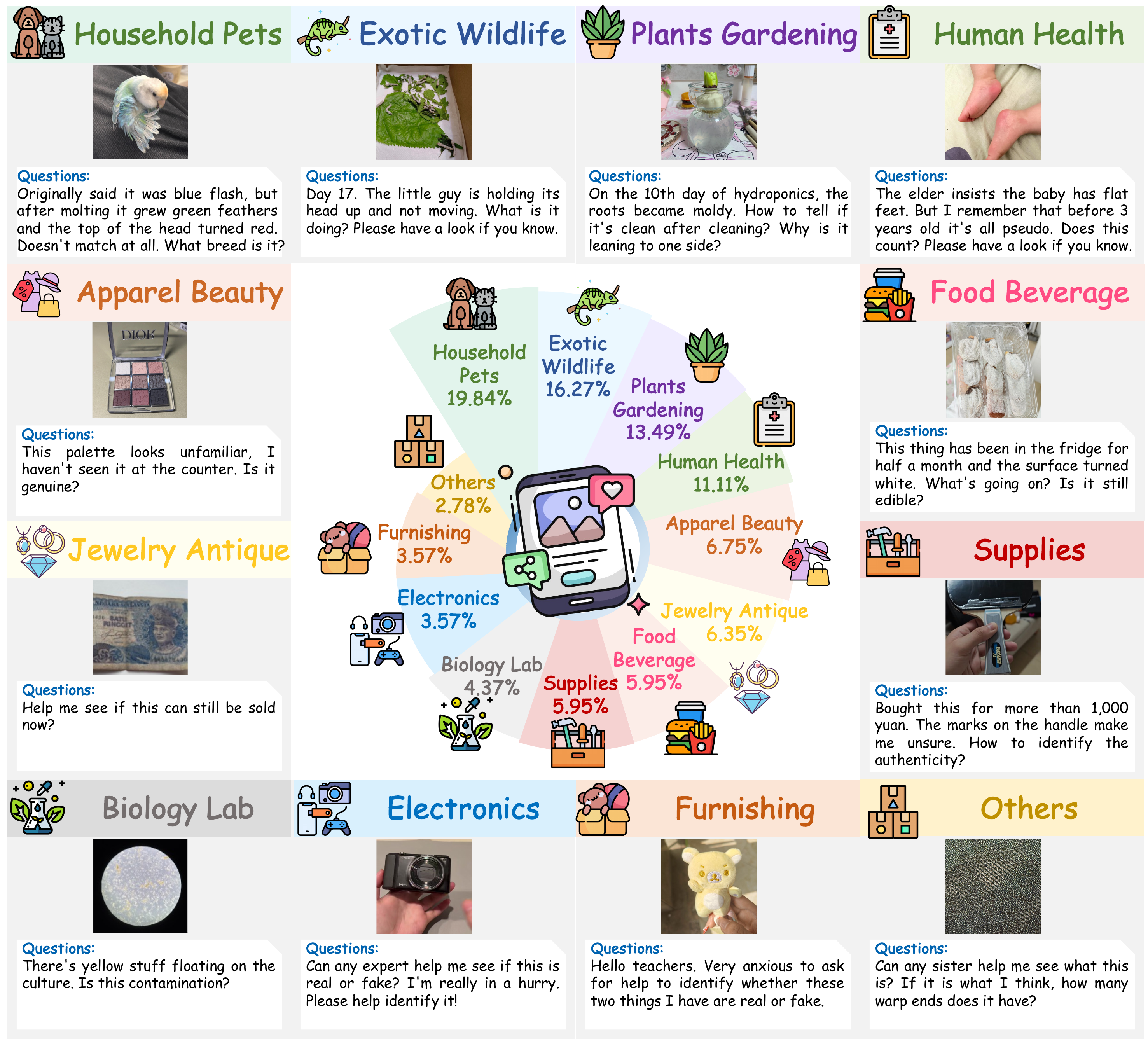}
\caption{Dataset composition and representative examples from NoteVQA. The figure shows the distribution across 12 topical categories with representative real-user query instances for each category.}
\label{fig:dataset-comp}
\end{figure}

The resulting NoteVQA benchmark contains 252 items spanning 12 categories and 7 intents. Queries average 22.7 words, while short references and interleaved multimodal summaries average 107.5 words and 219.4 words, respectively. The interleaved summaries contain 378 embedded images in total, averaging 1.5 images per item. Figure~\ref{fig:dataset-comp} shows the category distribution and representative queries. Full distributional statistics are provided in Appendix~\ref{app:stats-full}.

\subsection{Generating and Evaluating Interleaved Multimodal Answers}
\label{sec:framework-and-eval}

We provide two complementary components for the interleaved-answer
track: \textbf{AgenticInterleave} generates answers by retrieving
and integrating textual and visual evidence, while \textbf{IVR-12}
(\emph{Interleaved Visual-answer Rubric}) evaluates human-audited
references and model outputs on a common scale.

\paragraph{AgenticInterleave.}
Given a query--image pair, AgenticInterleave follows a single-agent
ReAct~\cite{yao2022react} loop with four retrieval tools:
\emph{text search}, \emph{page visit}, \emph{reverse image search},
and \emph{forward image search}. At each turn, the agent uses the
input and accumulated tool observations to decide whether to
retrieve further evidence or produce the final interleaved answer.
The loop ends when an answer is generated or repeated behaviour
is detected. In our experiments, the agent uses
Qwen3.5-397B-A17B as its backbone and receives only the query and
image; the short reference is used separately for reference
construction and content evaluation.

The system prompt guides image retrieval and placement through a
\textsc{Subject}/\textsc{Background} entity protocol.
\textsc{Subject} entities are the primary objects or concepts
addressed by the question, while \textsc{Background} entities
provide contextual, comparative, or supporting information.
When illustrating a \textsc{Subject} entity, the agent must
explicitly retrieve its image through forward image search.
The agent selects images directly from retrieval results without
a separate image-selection model and places each image tag
immediately after the sentence describing the corresponding entity.
Each tag records the forward image-search query and the index of
a returned candidate, which a post-hoc validator checks against
the tool results to verify retrieval provenance.
Appendix~\ref{app:framework-prompt} provides the system prompt,
tool definitions, and image-tag specification.

\paragraph{IVR-12: Interleaved Answer Evaluation.}
We assess answer correctness and interleaved-answer quality
separately. Short-answer accuracy is the proportion of responses
classified as correct by a SimpleQA-style judge. IVR-12 complements
this metric by evaluating the content, presentation, and images
of an interleaved answer, using identical criteria for
human-audited references and AgenticInterleave outputs.

IVR-12 comprises 12 dimensions scored from 1 to 5.
The five \emph{content} dimensions assess intent understanding,
objective accuracy, source fidelity, completeness, and neutrality
and safety. The three \emph{presentation} dimensions assess
reading experience, formatting and highlighting, and conciseness.
The four \emph{image} dimensions assess relevance, necessity and
placement, visual quality, and non-redundancy.
Three additional pass/fail gates cover timeliness, safety redlines,
and refusal correctness. Let $\bar{C}$, $\bar{P}$, and $\bar{I}$
denote the mean scores within the three categories. The overall
score is
\[
S_{\mathrm{IVR\text{-}12}} =
\begin{cases}
0.50\,\bar{C} + 0.20\,\bar{P} + 0.30\,\bar{I},
& \text{if all three gates pass},\\
0,
& \text{otherwise}.
\end{cases}
\]
Thus, answers that pass all gates receive a weighted score, while failure on any gate yields zero.

We use Qwen3.5-397B-A17B for both correctness judging and IVR-12
evaluation. For IVR-12, content and presentation are assessed
from the answer text, with the human-audited short reference
additionally supplied for content scoring. Image assessment
also uses the query image and all embedded images, while gate
checks use the query image and answer text.
Appendix~\ref{app:rubric-criteria} details the scoring criteria,
and Appendix~\ref{app:grader-prompts} provides the judge prompts,
parsing rules, and missing-value handling.

\section{Experiments}
\label{sec:exp}

We evaluate short-answer accuracy using nine models in thinking mode: Gemini-3.1-Pro~\cite{gemini31pro}, Doubao-Seed-2.0-Pro~\cite{seed2026seed2}, GLM-5.3-Flash~\cite{glm-5.3-flash}, Kimi-K2.6~\cite{kimi_k26}, Kimi-K3~\cite{team2026kimi}, Qwen3.8-Flash-Next~\cite{qwen3.8-flash}, Claude Opus 4.7~\cite{claude-opus-4.7}, Claude Sonnet 4.6~\cite{claude-sonnet-4-6} and Qwen3.5-397B-A17B~\cite{team2026qwen3}.
For interleaved-answer quality, we use IVR-12 to compare the outputs of Qwen3.5-397B-A17B running AgenticInterleave with the human-audited references.
Qwen3.5-397B-A17B is the default backbone for AgenticInterleave. It uses the four retrieval tools described in Section~\ref{sec:framework-and-eval}, with Serper~\footnote{\url{https://serper.dev/}} as the backend. The same model also serves as the IVR-12 judge.

\subsection{Main Results}
\label{subsec:main_results}

Table~\ref{tab:main} reports short-answer accuracy overall and by category; Table~\ref{tab:l2-intent} (Appendix~\ref{app:l2}) reports accuracy by intent. Gemini-3.1-Pro achieves the highest overall accuracy at 52.8\%. Adding agentic retrieval to Qwen3.5-397B-A17B raises its accuracy from 35.7\% to 37.7\%, a gain of 2.0\%.

\begin{table}[ht]
\centering
\color{abyss}
\caption{NoteVQA short-answer accuracy (\%). All models use thinking mode; \(^{\ddagger}\) denotes agentic retrieval.}
\label{tab:main}
\setlength{\tabcolsep}{4pt}
\renewcommand{\arraystretch}{1.10}
\setlength{\aboverulesep}{0pt}
\setlength{\belowrulesep}{0pt}

\fontsize{9}{11}\selectfont
\begin{tabular}{l *{13}{r}}
\toprule
\text{Model} & \text{Overall} & \text{Pet} & \text{Exo} & \text{Pln} &
\text{Hlt} & \text{Fas} & \text{Jwl} & \text{App} & \text{Fud} &
\text{Bio} & \text{CE} & \text{Hom} & \text{Oth} \\
\midrule
Gemini-3.1-Pro
& 52.8 & 50.0 & 46.3 & 61.8 & 64.3 & 58.8 & 37.5 & 60.0 & 60.0 & 45.5 & 55.6 & 55.6 & 14.3 \\
Doubao-seed-2.0-pro
& 49.6 & 48.0 & 53.7 & 47.1 & 64.3 & 41.2 & 18.8 & 73.3 & 53.3 & 45.5 & 55.6 & 55.6 & 14.3 \\
GLM-5.3-Flash
& 48.0 & 46.0 & 48.8 & 52.9 & 53.6 & 23.5 & 43.8 & 53.3 & 53.3 & 45.5 & 55.6 & 44.4 & 57.1 \\
Kimi-K2.6
& 47.6 & 42.0 & 43.9 & 50.0 & 57.1 & 52.9 & 56.2 & 60.0 & 46.7 & 36.4 & 44.4 & 44.4 & 28.6 \\
Qwen3.8-Flash-Next
& 44.0 & 42.0 & 46.3 & 58.8 & 50.0 & 29.4 & 43.8 & 46.7 & 33.3 & 45.5 & 33.3 & 22.2 & 42.9 \\
Claude-Opus-4.7
& 43.7 & 42.0 & 41.5 & 61.8 & 50.0 & 41.2 & 31.2 & 60.0 & 46.7 & 36.4 & 11.1 & 22.2 & 28.6 \\
Kimi-K3
& 43.3 & 38.0 & 41.5 & 52.9 & 57.1 & 41.2 & 31.2 & 60.0 & 46.7 & 36.4 & 22.2 & 44.4 & 14.3 \\
Claude-Sonnet-4.6
& 35.3 & 22.0 & 41.5 & 35.3 & 60.7 & 23.5 & 12.5 & 40.0 & 33.3 & 45.5 & 33.3 & 33.3 & 57.1 \\
Qwen3.5-397B-A17B
& 35.7 & 28.0 & 31.7 & 38.2 & 42.9 & 47.1 & 25.0 & 53.3 & 40.0 & 27.3 & 33.3 & 33.3 & 42.9 \\
Qwen3.5-397B-A17B$^{\ddagger}$
& 37.7 & 22.0 & 34.1 & 47.1 & 50.0 & 35.3 & 37.5 & 46.7 & 53.3 & 36.4 & 44.4 & 33.3 & 28.6 \\
\bottomrule
\end{tabular}
\par\smallskip
\begin{minipage}{\linewidth}
\footnotesize\raggedright
\textit{Categories:} Pet, HomePet; Exo, Exotic and Wild; Pln, Plant and Garden; Hlt, Human Health; Fas, Fashion and Beauty; Jwl, Jewellery and Antique; App, Appliance, Kitchen, and Car; Fud, Food and Beverage; Bio, BioLab; CE, Consumer Electronics; Hom, Home; Oth, Other.
\end{minipage}
\end{table}

Table~\ref{tab:main-ivr} compares interleaved-answer quality. The references score 4.65 under IVR-12, compared with 3.52 for AgenticInterleave, a gap of 1.13 points.

The references score higher on all 12 dimensions.
The largest differences concern Content, whose mean score is 4.82 for the references and 2.95 for AgenticInterleave. Accuracy shows the largest gap (2.68 points), followed by source fidelity (2.07), completeness (1.90), and intent understanding (1.76).
Scores for Presentation and Image are closer to those of the references, with mean gaps of 0.37 and 0.40, respectively. Formatting shows the largest gap within these two categories (0.86). Thus, the main shortfall under IVR-12 lies in answer content.

\begin{table}[H]
\centering
\color{abyss}
\caption{Mean IVR-12 scores for human-audited references and AgenticInterleave outputs.}
\label{tab:main-ivr}

\setlength{\tabcolsep}{4pt}
\renewcommand{\arraystretch}{1.10}
\setlength{\aboverulesep}{0pt}\setlength{\belowrulesep}{0pt}

\fontsize{9}{11}\selectfont
\begin{tabular}{l r r r r r r r r r r r r r}
\toprule
\multirow{2}{*}{System} & \multirow{2}{*}{$S_{\mathrm{IVR-12}}$} & \multicolumn{5}{c}{Content} & \multicolumn{3}{c}{Presentation} & \multicolumn{4}{c}{Image} \\
\cmidrule(lr){3-7}\cmidrule(lr){8-10}\cmidrule(lr){11-14}
 &  & Int & Acc & Src & Cmp & Sfy & Rd & Fmt & Cnc & Rel & Pos & Qty & NR \\
\midrule
Human-audited gold  & 4.65 & 4.94 & 4.90 & 4.48 & 4.84 & 4.92 & 5.00 & 4.96 & 4.81 & 4.17 & 4.16 & 4.24 & 4.18 \\
AgenticInterleave   & 3.52 & 3.18 & 2.22 & 2.41 & 2.94 & 4.00 & 4.94 & 4.10 & 4.63 & 3.79 & 3.79 & 3.81 & 3.76 \\
\bottomrule
\end{tabular}
\par\smallskip
\begin{minipage}{\linewidth}
\footnotesize\raggedright
\textit{Dimensions:} Int, intent understanding; Acc, accuracy; Src, source fidelity; Cmp, completeness; Sfy, neutrality and safety; Rd, reading experience; Fmt, formatting; Cnc, conciseness; Rel, image relevance; Pos, necessity and position; Qty, image quality; NR, non-redundancy.
\end{minipage}
\end{table}

\subsection{Analysis of Agentic Search Gains}
\label{subsec:attribution}

To understand the limited aggregate gain from agentic retrieval, we group all 252 items by whether the reasoning-only Qwen3.5-397B-A17B baseline and its agentic counterpart answer correctly.
Table~\ref{tab:buckets} reports the four groups and their tool usage. Retrieval corrects 33 baseline errors but introduces 28 new errors, leaving a net gain of five correct answers. Both settings fail on 129 items (51.2\%).

\begin{table}[H]
\centering

\color{abyss}
\setlength{\tabcolsep}{4pt}
\renewcommand{\arraystretch}{1.18}
\setlength{\aboverulesep}{0pt}
\setlength{\belowrulesep}{0pt}

\caption{Qwen3.5-397B-A17B answer outcomes with and without retrieval, and tool usage in agentic runs. 
\(\surd\) and \(\times\) denote correct and incorrect answers, respectively.}
\label{tab:buckets}
\fontsize{9}{11}\selectfont
\begin{tabular}{l c c r r r r r r}
\toprule
\text{Bucket}
& \text{Base}
& \text{Agent}
& \text{N}
& \text{Share (\%)}
& \text{Rounds}
& \text{Search}
& \text{Visit}
& \text{Img.} \\
\midrule
Both wrong
& $\times$
& $\times$
& 129 & 51.2 & 4.60 & 6.48 & 1.25 & 0.69 \\
Both correct
& $\surd$
& $\surd$
& 62 & 24.6 & 4.31 & 5.81 & 1.34 & 0.55 \\
Search helps
& $\times$
& $\surd$
& 33 & 13.1 & 4.73 & 6.91 & 1.24 & 0.70 \\
Search hurts
& $\surd$
& $\times$
& 28 & 11.1 & 4.89 & 7.11 & 1.75 & 0.54 \\
\midrule
Total
& -- & --
& 252 & 100.0 & 4.58 & 6.44 & 1.33 & 0.64 \\
\bottomrule
\end{tabular}
\par\smallskip
\begin{minipage}{\linewidth}
\footnotesize\raggedright
Base: reasoning-only; Agent: agentic retrieval. Rounds, Search, Visit, and Img. report mean counts per agentic run; Img. denotes image search.
\end{minipage}
\end{table}

Correct and incorrect agentic runs use similar numbers of tool calls on average: 6.19 vs. 6.59 web searches, 1.31 vs. 1.34 page visits, and 0.60 vs. 0.66 image searches. The \emph{Search helps} group averages 6.91 searches, compared with 7.11 for \emph{Search hurts}. Useful-evidence hit rates are 0.12\% across 1623 web-search calls and 5.69\% across 334 page visits.
To examine the failure patterns, a separate Qwen3.5-397B-A17B judge classifies the 157 incorrect agentic runs into tool-related (T1--T4) and non-tool (N1--N2) categories (Table~\ref{tab:errors}).

\begin{table}[!ht]
\centering
\color{abyss}

\setlength{\tabcolsep}{4pt}
\renewcommand{\arraystretch}{1.18}
\setlength{\aboverulesep}{0pt}
\setlength{\belowrulesep}{0pt}

\caption{Judge-assigned failure categories for 157 incorrect agentic retrieval runs. 
}
\label{tab:errors}
\vspace{4pt}

\fontsize{9}{11}\selectfont
\begin{tabular}{l l r r r r r r r}
\toprule
\text{ID} &
\text{Failure cause} &
\text{BW} &
\text{Loss} &
\text{Total} &
\text{Share (\%)} &
\text{Rounds} &
\text{Search} &
\text{Img.} \\
\midrule
\multicolumn{9}{l}{\textit{Tool-related failures}} \\
T1 & Reverse image search ineffective
& 8 & 2 & 10 & 6.4 & 5.30 & 6.70 & 1.00 \\
T2 & Insufficient text-search coverage
& 9 & 2 & 11 & 7.0 & 6.36 & 7.27 & 0.91 \\
T3 & Confirmation / anchoring loop
& 90 & 21 & 111 & 70.7 & 4.56 & 6.96 & 0.64 \\
T4 & Cross-lingual query dilution
& 3 & 1 & 4 & 2.5 & 4.50 & 7.00 & 0.75 \\
\midrule
\multicolumn{9}{l}{\textit{Non-tool failures}} \\
N1 & Subjective reference or strict judging
& 13 & 2 & 15 & 9.6 & 4.13 & 4.47 & 0.53 \\
N2 & Other reasoning or execution errors
& 6 & 0 & 6 & 3.8 & 3.67 & 3.33 & 0.33 \\
\midrule
\text{Total} & ---
& 129 & 28 & 157 & 100.0 & --- & --- & --- \\
\bottomrule
\end{tabular}
\par\smallskip
\begin{minipage}{\linewidth}
\footnotesize\raggedright
BW: both wrong; Loss: Search hurts. Rounds, Search, and Img. are mean counts per run; Img. denotes image search.
\end{minipage}
\end{table}

Under this classification, tool-related failures account for 136 of the 157 incorrect runs (86.6\%). Confirmation or anchoring loops (T3) are the largest category, with 111 cases (70.7\%).
In these trajectories, the agent often adopts an incorrect hypothesis early and uses subsequent searches to reinforce it, without seeking evidence that would challenge it.
The \emph{Search hurts} group also averages more web-search calls than the \emph{Both wrong} group (7.11 vs. 6.48). This pattern is consistent with continued retrieval along an incorrect search direction, although call counts alone do not establish the cause of failure.
These results suggest two obstacles to effective retrieval: obtaining relevant evidence from Chinese UGC and niche expert communities, and revising an incorrect hypothesis once search is under way.
Together, the paired outcomes and failure analysis show why the evaluated agentic configuration yields only a small net improvement on NoteVQA.

\subsection{Image Ablation of Interleaved References}
\label{sec:img-ablation}

We assess the contribution of embedded images using probes derived from the 205 references that contain at least one image. A multimodal generator creates four probes per item: two \emph{image-critical} probes, designed to require information from an image, and two \emph{text-answerable} probes, designed to be answerable from the paragraph alone. A positive-control filter removes probes that the reader cannot answer with the full interleaved input, retaining 367 image-critical and 375 text-answerable probes.

The reader, Qwen3.5-397B-A17B-FP8, answers each probe with the same paragraph text under three conditions: image tags removed ($y_\text{text}$), tags replaced by \texttt{[image N]} placeholders ($y_\text{ph}$), or images displayed inline ($y_\text{full}$). The reader receives neither the original query image nor the probe reference answer. A text-only judge grades each response. We report 95\% confidence intervals from 2000 bootstrap resamples.

\begin{table}[H]
\color{abyss}
\setlength{\aboverulesep}{0pt}\setlength{\belowrulesep}{0pt}
\begin{minipage}[t]{0.49\textwidth}
\centering

\setlength{\tabcolsep}{4pt}
\renewcommand{\arraystretch}{1.10}
\caption{Reader accuracy on retained probes under image ablation (\%; 95\% bootstrap CIs).}
\label{tab:ablation}
\vspace{3pt}
\fontsize{9}{11}\selectfont
\begin{tabular}{l l r r r}
\toprule
Probe type & Cond. & Hits/N & Acc \% & 95\% CI \\
\midrule
Image-crit. & $y_\text{text}$ & 76/367  & 20.7 & [16.6, 25.1] \\
Image-crit. & $y_\text{ph}$   & 115/367 & 31.3 & [26.7, 36.2] \\
Image-crit. & $y_\text{full}$ & 356/367 & 97.0 & [95.4, 98.6] \\
\midrule
Text-answ. & $y_\text{text}$ & 369/375 & 98.4 & [97.1, 99.5] \\
Text-answ. & $y_\text{ph}$   & 367/375 & 97.9 & [96.3, 99.2] \\
Text-answ. & $y_\text{full}$ & 368/375 & 98.1 & [96.5, 99.5] \\
\bottomrule
\end{tabular}
\end{minipage}%
\hfill
\begin{minipage}[t]{0.49\textwidth}
\centering

\setlength{\tabcolsep}{4pt}
\renewcommand{\arraystretch}{1.10}
\caption{Image-critical accuracy by intent (\%). $\Delta_\text{f-t}$: full minus text (\%).}
\label{tab:ablation-l2}
\vspace{3pt}
\fontsize{9}{11}\selectfont
\begin{tabular}{l r r r r r}
\toprule
Intent & N & $y_\text{text}$ & $y_\text{ph}$ & $y_\text{full}$ & $\Delta_\text{f-t}$ \\
\midrule
Authenticity   & 66 & 13.6 & 19.7 &  98.5 & +84.8 \\
Diagnosis      & 65 & 30.8 & 36.9 & 100.0 & +69.2 \\
Identification & 53 & 22.6 & 30.2 &  98.1 & +75.5 \\
Object cond.   & 35 &  8.6 & 28.6 &  88.6 & +80.0 \\
Physiology     & 57 & 28.1 & 40.4 &  94.7 & +66.7 \\
Procedural     & 49 & 12.2 & 26.5 &  98.0 & +85.7 \\
Spoilage       & 42 & 23.8 & 38.1 &  97.6 & +73.8 \\
\midrule
Overall        & 367 & 20.7 & 31.3 & 97.0 & +76.3 \\
\bottomrule
\end{tabular}
\end{minipage}
\end{table}

Table~\ref{tab:ablation} shows that image-critical accuracy rises from 20.7\% with text alone to 97.0\% with the full interleaved input, a gain of 76.3\%. Placeholder markers raise accuracy to 31.3\%, a gain of 10.6\% over text alone; displaying the images adds a further 65.7\%. Text-answerable accuracy remains between 97.9\% and 98.4\%, varying by at most 0.5\% across conditions.

The full-minus-text accuracy gain is positive across all seven intents, ranging from +66.7 to +85.7\% (Table~\ref{tab:ablation-l2}). In the 281 probes corrected by $y_\text{full}$, the text-only responses report missing information that the images subsequently provide. These results indicate that embedded images supply useful information beyond the paragraph text on the selected image-critical probes.

\section{Related Work}
\label{sec:related_work}

\paragraph{Multimodal QA benchmarks.}
Existing multimodal QA benchmarks evaluate factual knowledge, visual understanding, and information seeking through answer-level correctness. SimpleVQA~\cite{cheng2025simplevqa} and FVQA~\cite{wang2017fvqa} focus on factual visual question answering, while LiveVQA~\cite{fu2025livevqa} examines the acquisition and use of up-to-date visual knowledge. MM-BrowseComp~\cite{li2025mm}, MMSearch~\cite{jiang2024mmsearch}, and MMSearch-Plus~\cite{tao2026mmsearch} extend evaluation to multimodal browsing and retrieval, with MMSearch-Plus emphasizing fine-grained visual cues and provenance verification. BC-VL~\cite{geng2026webwatcher} and VDR-Bench~\cite{zeng2026vision} further construct challenging visual search tasks through entity fuzzification and knowledge-graph-based query expansion, respectively, reducing opportunities for text-only shortcuts or simple image matching. These benchmarks advance the evaluation of visual reasoning and search, but their answer-level metrics do not directly assess the quality of a complete explanation supported by embedded images. NoteVQA complements this line of work with community-sourced visual questions and paired short and interleaved references, enabling both correctness and explanation-quality evaluation.

\paragraph{Multimodal deep research.}
Recent work extends multimodal evaluation from factual answers to long-form research reports. MMDR-Bench~\cite{huang2026mmdeepresearch} comprises 140 expert-crafted tasks across 21 domains and evaluates multimodal understanding, citation grounding, and report synthesis using image--text task bundles. MiroEval~\cite{ye2026miroeval} contains 100 tasks, including 30 multimodal tasks, constructed from real user query patterns and web trends; its evaluation covers synthesis quality, factuality, and the research process. Complementing these benchmarks, PTAH~\cite{zhang2026towards} coordinates planning, research, writing, and verification agents to generate interleaved reports, with additional evaluation of image content and multimodal presentation quality. These efforts primarily target extended research workflows and report generation. NoteVQA focuses on everyday visual questions posted in online communities and provides human-audited interleaved references alongside short answers, connecting factual QA evaluation with the quality of visually supported responses.

\paragraph{Rubric design and model-based evaluation.}
Multimodal report evaluation increasingly considers both textual claims and their visual support. MMDR-Bench~\cite{huang2026mmdeepresearch} evaluates visual evidence fidelity and the consistency between image-referenced claims and their associated visual content. PTAH~\cite{zhang2026towards} assesses image content and rendered presentation through dimensions including cross-modal alignment, information complementarity, and density-legibility balance. Building on these directions, IVR-12 evaluates interleaved answers through 12 dimensions grouped into content, presentation, and image quality. Its image dimensions separately assess relevance, necessity and placement, visual quality, and non-redundancy, while its content and presentation dimensions capture whether an answer is accurate, useful, and readable. We apply the same rubric to human-audited references and model outputs, providing a common scale for comparing interleaved-answer quality on real-world visual questions.

\section{Conclusion}

We introduce NoteVQA, a benchmark of 252 community-sourced visual
questions that connects real user needs with evaluations of answer
correctness and visually grounded explanation quality. Each question
is paired with a short reference and a human-audited interleaved
reference, complemented by AgenticInterleave for answer generation
and IVR-12 for evaluation. Across 9 frontier VLMs, the highest
short-answer accuracy is only 52.8\%, highlighting the substantial
challenge posed by everyday visual questions. Agentic retrieval
improves Qwen3.5-397B-A17B by just 2.0 percentage points, while its
interleaved answers score 3.52 under IVR-12 compared with 4.65 for
the references, with the largest gap in content quality. In the
image-ablation study, including embedded images raises reader
accuracy on image-critical probes from 20.7\% to 97.0\%, while
performance on text-answerable controls remains stable. This
demonstrates that the images contribute information absent from
the accompanying text. Together, these findings highlight the
need for VLMs that can both answer real-world visual questions
accurately and communicate the supporting evidence effectively.
We will release the benchmark, evaluation prompts, model outputs,
and code for the generation framework and data construction pipeline.

\section*{Contributors}
\begin{flushleft}
\small
\begingroup
\renewcommand{\thefootnote}{*}
\hypersetup{linkcolor=abyss}
Haonan Jiang, Guojian Zhan, Jiancong Xie, Shijun Wan, Dongiia Zhao, Cheng Chen, Yahui Liu\emailicon, Chuan Mu\emailicon
\endgroup
\end{flushleft}

\clearpage

\bibliographystyle{styles/colm2026_conference}
\bibliography{bibliography/references}

\clearpage

\appendix

\section{Full Benchmark Statistics}
\label{app:stats-full}

\begin{table}[H]
\centering\small\color{abyss}
\setlength{\tabcolsep}{5pt}
\renewcommand{\arraystretch}{1.12}
\caption{Full statistics of NoteVQA on the Stage-11 human-reviewed set.}
\label{tab:stats-full}
\vspace{4pt}
\setlength{\aboverulesep}{0pt}\setlength{\belowrulesep}{0pt}
\begin{tabular}{l r}
\toprule
Statistic & Value \\
\midrule
Number of items $N$                                        & 252 \\
Categories                                              & 12 \\
Intents                                                 & 7 \\
Query length in words, mean $\pm$ std, median              & $22.7 \pm 11.0$, $22.0$ \\
Short-answer length in words, mean $\pm$ std, median       & $107.5 \pm 40.2$, $102.5$ \\
Interleaved answer length in words, mean $\pm$ std, median & $219.4 \pm 47.8$, $215.0$ \\
Total embedded images                                      & 378 \\
Images per item, mean $\pm$ std, median                    & $1.5 \pm 1.1$, $1.0$ \\
Embed-count distribution over 0, 1, 2, 3, and 4+ images    & 47 / 89 / 74 / 32 / 10 \\
\bottomrule
\end{tabular}
\end{table}

\begin{table}[H]
\centering\small\color{abyss}
\setlength{\tabcolsep}{4pt}
\renewcommand{\arraystretch}{1.10}
\caption{Category and intent distribution matrix. A dot marks an empty cell.}
\label{tab:cat-intent}
\vspace{4pt}
\setlength{\aboverulesep}{0pt}\setlength{\belowrulesep}{0pt}
\resizebox{\linewidth}{!}{%
\begin{tabular}{l r r r r r r r r}
\toprule
Category $\backslash$ Intent & Diagnosis & Authenticity & Physiology & Identification & Procedural & Object Condition & Spoilage & Total \\
\midrule
HomePet     & 17 &  3 & 16 &  4 &  3 &  3 &  4 & 50 \\
Exotic      &  6 &  3 &  9 & 19 &  2 &  1 &  1 & 41 \\
Plant       &  8 &  3 &  6 &  6 & 10 & $\cdot$ &  1 & 34 \\
Health      & 14 &  5 &  8 & $\cdot$ &  1 & $\cdot$ & $\cdot$ & 28 \\
Fashion     &  2 &  8 & $\cdot$ & $\cdot$ &  1 &  6 & $\cdot$ & 17 \\
Jewelry     & $\cdot$ & 12 & $\cdot$ &  1 & $\cdot$ &  2 &  1 & 16 \\
Food        & $\cdot$ & $\cdot$ & $\cdot$ &  1 & $\cdot$ & $\cdot$ & 14 & 15 \\
Appliance   & $\cdot$ &  3 & $\cdot$ & $\cdot$ &  4 &  8 & $\cdot$ & 15 \\
BioLab      &  3 & $\cdot$ &  4 & $\cdot$ &  1 &  1 &  2 & 11 \\
Home        & $\cdot$ &  3 & $\cdot$ &  1 &  1 &  3 &  1 & 9 \\
CE          & $\cdot$ &  4 & $\cdot$ & $\cdot$ &  3 &  2 & $\cdot$ & 9 \\
Other       & $\cdot$ & $\cdot$ & $\cdot$ &  4 &  3 & $\cdot$ & $\cdot$ & 7 \\
\midrule
Total & 50 & 44 & 43 & 36 & 29 & 26 & 24 & 252 \\
\bottomrule
\end{tabular}}
\end{table}

\section{Intent Accuracy Full Grid}
\label{app:l2}

\begin{table}[H]
\centering\small\color{abyss}
\setlength{\tabcolsep}{5pt}
\renewcommand{\arraystretch}{1.12}
\caption{Short‑answer accuracy across 7 intents. All models evaluated in thinking mode. Rows marked \(^{*}\) use the four‑tool agentic retrieval suite of Section~\ref{sec:framework-and-eval}; others are single‑pass no‑tool.}
\label{tab:l2-intent}
\vspace{4pt}
\setlength{\aboverulesep}{0pt}\setlength{\belowrulesep}{0pt}
\resizebox{\linewidth}{!}{%
\begin{tabular}{l r r r r r r r r}
\toprule
Model & Overall & Diagnosis & Authenticity & Physiology & Identification & Procedural & Object Condition & Spoilage \\
\midrule
Gemini-3.1-Pro                    & 52.8 & 60.0 & 38.6 & 67.4 & 27.8 & 65.5 & 57.7 & 54.2 \\
Doubao-seed-2.0-pro               & 49.6 & 66.0 & 29.5 & 65.1 & 30.6 & 62.1 & 50.0 & 37.5 \\
GLM-5.3-Flash                     & 48.0 & 52.0 & 27.3 & 55.8 & 36.1 & 86.2 & 30.8 & 54.2 \\
Kimi-K2.6                         & 47.6 & 52.0 & 36.4 & 51.2 & 33.3 & 62.1 & 57.7 & 45.8 \\
Qwen3.8-Flash                     & 44.0 & 60.0 & 31.8 & 55.8 & 44.4 & 51.7 & 23.1 & 25.0 \\
Claude-4.7-Opus                   & 43.7 & 58.0 & 29.5 & 48.8 & 25.0 & 62.1 & 42.3 & 37.5 \\
Kimi-K3                           & 43.3 & 56.0 & 27.3 & 46.5 & 25.0 & 55.2 & 50.0 & 45.8 \\
Claude-4.6-Sonnet                 & 35.3 & 44.0 & 20.5 & 34.9 & 27.8 & 48.3 & 34.6 & 41.7 \\
Qwen3.5-397B-A17B                 & 35.7 & 40.0 & 27.3 & 37.2 & 25.0 & 51.7 & 30.8 & 41.7 \\
Qwen3.5-397B-A17B$^{*}$   & 37.7 & 46.0 & 31.8 & 32.6 & 27.8 & 37.9 & 38.5 & 54.2 \\
\bottomrule
\end{tabular}}
\end{table}

Excluding the agentic row, Doubao-seed-2.0-pro leads Diagnosis at 66.0\%, Gemini-3.1-Pro leads Authenticity at 38.6\%, Qwen3.8-Flash leads Identification at 44.4\%, and GLM-5.3-Flash leads Procedural at 86.2\%. Gemini-3.1-Pro tie for the highest Physiology accuracy at 67.4\%; Gemini-3.1-Pro and Kimi-K2.6 tie for Object Condition at 57.7\%; and Gemini-3.1-Pro and GLM-5.3-Flash tie for Spoilage at 54.2\%. Across all reported configurations, Authenticity is the only intent whose best accuracy remains below 40\%, at 38.6\%.

\section{Interleaved Multimodal Summary Construction Pipeline Details}
\label{app:pipeline-full}

This appendix details the pipeline behind the human-verified interleaved multimodal summary described in Section~\ref{sec:interleaved-gt} and Figure~\ref{fig:interleaved-pipeline}: how the per-item image evidence pool is built from an agent trajectory, and how that pool is then synthesised into an interleaved answer and cleaned. For each stage we report the prompts, thresholds, and model-call parameters.

\subsection{Evidence Pool Construction}

\paragraph{Contributing URL Ranking.} A rule-based pre-pass processes each agent episode and emits one entry per tool result, covering three tool types: search results, image search results with the returned image URL preserved, and visited pages where each visited URL becomes its own entry split by the boundary pattern ``The useful information in url for user goal $\ldots$ as follows:''. The complete URL list for a task, aggregated across all three tool types, is then packaged into a single text-only prompt whose fields are the question, the reference answer, and one \mbox{[ID $i$] type=$t$ | URL: $u$ $\backslash$n $c$} block per URL. Qwen scores each URL on relevance, completeness, clarity, answer-support, and necessity; the five scores are summed and the top-5 URLs per item are retained. 

\begin{promptbox}{URL Contribution Scoring}
\begin{lstlisting}[style=promptstyle]
You are evaluating which web pages contributed most to correctly answering a research question.

**Question**: {query}
**Expected Answer**: {answer}

Below are all web pages collected during research, each with an ID, type, URL, and content:

{links_block}

---

Score each page on 5 dimensions (1-5) and select the TOP {top_k} most useful pages.
Return ONLY valid JSON with the top {top_k} page IDs. Be concise in the reason field.

Output format:
{
  "<id>": {
    "reason": "<one-sentence explanation>",
    "relevance_score": <1-5>,
    "completeness_score": <1-5>,
    "clarity_score": <1-5>,
    "answer_support_score": <1-5>,
    "necessity_score": <1-5>
  },
  ...
}

Scoring criteria:
- relevance_score (1-5): Is the page content directly relevant to the question?
- completeness_score (1-5): Does the page contain effective information for answering?
- clarity_score (1-5): Is the content clear, well-structured, and readable?
- answer_support_score (1-5): Does the page support or verify the expected answer?
- necessity_score (1-5): Is the page essential (not just supplementary)?
\end{lstlisting}
\end{promptbox}

\paragraph{Web-Page Image Crawling.} Each of the top-5 URLs is fetched through Serper's \texttt{scrape} endpoint with \texttt{includeMarkdown=true}; the returned \texttt{og:image} is prepended to the markdown so hero images are not lost. Candidate image URLs are then harvested with the union of the three regexes in Table~\ref{tab:image-url-patterns}, and any URL matching the case-insensitive ad/tracker blocklist in Table~\ref{tab:blocklist} is dropped before download. We attempt to download up to $60$ candidate images per URL; we retain the first $20$ whose shorter side is at least $100$\,pixels (\texttt{MIN\_PX}$=100$, checked via \texttt{PIL.Image.size}). The download timeout per image is $10\,$s with a browser \texttt{User-Agent} and concurrency $32$. Because the previous stage rarely selects an \texttt{image\_search} URL --- its textual \texttt{content} is short --- a parallel path additionally downloads every \texttt{image\_url} returned by \texttt{image\_search} calls (subject to the same $100$\,px floor and blocklist) into a separately-tracked \texttt{image\_search\_pool}, ensuring visually valuable results from reverse search are not discarded.

\begin{table}[H]
\centering
\caption{Image-URL extraction patterns used on the scraped markdown. The three regexes are applied independently and their captured URL sets are unioned.}
\label{tab:image-url-patterns}
\resizebox{\linewidth}{!}{%
\begin{tabular}{ll}
\toprule
Source & Regex (URL captured in group 1) \\
\midrule
Markdown       & \texttt{!\textbackslash[[\textasciicircum\textbackslash]]*\textbackslash]\textbackslash((https?://[\textasciicircum\textbackslash s)]+)\textbackslash)} \\
HTML \texttt{<img>} & \texttt{<img[\textasciicircum>]+src=["\textbackslash']?(https?://[\textasciicircum\textbackslash s"\textbackslash'><]+)} (case-insensitive) \\
CSS background & \texttt{background(?:-image)?\textbackslash s*:\textbackslash s*url\textbackslash(["\textbackslash']?(https?://[\textasciicircum\textbackslash s"\textbackslash')<]+)} (case-insensitive) \\
\bottomrule
\end{tabular}%
}
\end{table}

\begin{table}[H]
\centering
\caption{Ad/tracker/icon blocklist. A candidate image URL is dropped before download if it matches any of these substrings or path patterns (single case-insensitive regex; alternation shown here row-wise for readability).}
\label{tab:blocklist}
\begin{tabular}{ll}
\toprule
Category & Substring / pattern \\
\midrule
Ads             & \texttt{/ads?/}, \texttt{/banner}, \texttt{advertisement}, \texttt{doubleclick}, \texttt{googletagmanager} \\
Tracking pixels & \texttt{/pixel[./]}, \texttt{/tracking}, \texttt{/beacon}, \texttt{1x1}, \texttt{analytics}, \texttt{\textbackslash.gif\textbackslash?}, \texttt{facebook.com/tr} \\
Avatars / logos & \texttt{gravatar.com}, \texttt{/icon[s]?/}, \texttt{/logo.(png|svg|ico)\$}, \texttt{/favicon} \\
Chrome          & \texttt{/sprite}, \texttt{/placeholder} \\
\bottomrule
\end{tabular}
\end{table}

\begin{promptbox}{Candidate-Image Scoring}
\begin{lstlisting}[style=promptstyle]
You are an expert evaluator assessing the multi-dimensional contribution of images to answering questions.

## Input Information

Two images are provided above in order:
1. **Query Image** (first image): the original image attached to the question.
2. **Web Image** (second image): an image crawled from a contributing web page.

**Question:** {question}

**Expected Answer:** {answer}

## Evaluation Task

Evaluate how the **Web Image** (second image) contributes to answering the question, taking the Query Image as context.

1. **Visual Relevance** (1-5): Is the Web Image directly related to the question?
2. **Information Content** (1-5): Does the Web Image contain useful information to answer?
3. **Clarity** (1-5): Is the Web Image clear, readable, and well-presented?
4. **Answer Substantiation** (1-5): Does it help verify or clarify the expected answer?
5. **Necessity** (1-5): Is the Web Image essential or highly helpful vs optional?

## Output Format (JSON only)

Return ONLY valid JSON with no additional text:
{
  "visual_relevance": {"score": <1-5>, "reason": "<brief explanation>"},
  "information_content": {"score": <1-5>, "reason": "<brief explanation>"},
  "clarity": {"score": <1-5>, "reason": "<brief explanation>"},
  "answer_substantiation": {"score": <1-5>, "reason": "<brief explanation>"},
  "necessity": {"score": <1-5>, "reason": "<brief explanation>"},
  "overall_score": <1-5>
}
\end{lstlisting}
\end{promptbox}

\paragraph{Candidate-Image Scoring.} Every image in \texttt{best\_link[*].images} $\cup$ \texttt{image\_search\_pool} is scored by a two-image VLM call: the original query image is passed first, then the candidate. The two pools are then merged and stable-sorted by \texttt{overall\_score} in descending order; the first $10$ are retained, with their provenance recorded via a \texttt{source} tag ($\in$\{\texttt{best\_link},\,\texttt{image\_search\_pool}\}) so the two paths remain distinguishable downstream.

\subsection{Interleaved Multimodal Summary Synthesis}

\paragraph{Multimodal Summary Generation.} A single multimodal Qwen call receives, in order, the query image, up to ten candidate images (indexed \texttt{Image\,1..N}), and a text block containing the question, the reference answer, a formatted trajectory (each step's observation is capped at $3000$ characters), and the scores for each candidate laid out as in Table~\ref{tab:sub7-candidate-block}.  Tags emitted by the model must match the grammar \texttt{<image>\{"candidate\_images":\,N\}</image>}; after generation, a regex extracts every referenced index and \texttt{candidate\_images} is filtered down to those actually used. If no candidate image survived scoring for a given item, the generator receives a placeholder.

\vspace{6pt}

\begin{promptbox}{MMSummary Generator (System Prompt)}
\begin{lstlisting}[style=promptstyle]
You are an expert answer writer producing image-text interleaved (MMSummary) answers.

You will receive:
- a user QUESTION (with its query image, shown first),
- the CORRECT reference ANSWER,
- BACKGROUND research findings (tool call trajectory: search/visit/image_search results),
- a set of CANDIDATE IMAGES (each shown in order with an index Image 1, Image 2, ...).

Write a rich, detailed English answer that satisfies the following:

Style rules:
- Lead directly with the answer -- do NOT begin with "Based on..." / "According to..." / "The research shows...".
- Expand with specific facts, names, numbers, and context drawn from the background research.
- Write with authority, as if you already know the topic well.

Image embedding rules:
- Embed candidate images ONLY when they genuinely illustrate the neighbouring sentence.
- Prefer images with higher overall_score and stronger visual relevance to the sentence you just wrote.
- Insert tags in exactly this JSON format: <image>{"candidate_images": N}</image>
  where N is the 1-based Image index shown to you. Place the tag on its own line immediately after the sentence it illustrates.
- Do NOT invent image indices -- only use indices from the candidate images actually provided.
- If NO candidate image is a good fit, write a pure text answer without any <image> tags.
- Typical density: 1-4 image tags for a 200-400 word answer. Do not force insertion.

Return ONLY the answer text (with any embedded image tags). No preamble, no metadata.
\end{lstlisting}
\end{promptbox}

\begin{promptbox}{MMSummary Generator (User Prompt Template)}
\begin{lstlisting}[style=promptstyle]
## Question (Query image is the FIRST image shown above)
{question}

## Correct Reference Answer
{answer}

## Background Research (agent trajectory)
{trajectory_text}

## Candidate Images (shown AFTER the query image, in the order Image 1, Image 2, ...)
{format_candidate_scores(candidate_images)}

Now write the image-text interleaved MMSummary.
\end{lstlisting}
\end{promptbox}

\begin{table}[H]
\centering
\caption{Per-candidate score block passed to the generator (one such block for each of the up-to-10 images).}
\label{tab:sub7-candidate-block}
\begin{tabular}{ll}
\toprule
Line & Content \\
\midrule
1 & \texttt{Image \{idx\}\,\,(source=\{src\})\,\,reference tag: <image>\{"candidate\_images":\,\{idx\}\}</image>} \\
2 & \texttt{\,\,visual\_relevance:\,\{score\}\,--\,\{reason\}} \\
3 & \texttt{\,\,information\_content:\,\{score\}\,--\,\{reason\}} \\
4 & \texttt{\,\,clarity:\,\{score\}\,--\,\{reason\}} \\
5 & \texttt{\,\,answer\_substantiation:\,\{score\}\,--\,\{reason\}} \\
6 & \texttt{\,\,necessity:\,\{score\}\,--\,\{reason\}} \\
7 & \texttt{\,\,overall\_score:\,\{score\}} \\
\bottomrule
\end{tabular}
\end{table}

\paragraph{Five-Rule Filter and Cross-Image Deduplication.} Each embedded tag is re-examined by a two-image VLM call (query image + candidate) that decides whether to remove the candidate image using five rules --- \textsc{redundant-with-original}, \textsc{composite/collage}, \textsc{global-watermark}, \textsc{video-thumbnail} (a play-button overlay), and \textsc{off-topic}. A $\pm200$-character window around the tag is passed as \texttt{context} (truncated to $300$ characters); an API failure defaults to \emph{keep} and an image that cannot be loaded defaults to \emph{remove}. Images that survive the five-rule filter are then compared pairwise against every previously retained image via a single-question VLM probe that returns \texttt{\{"similar": true\}} to trigger deduplication, so near-duplicates within a summary collapse to one representative image.

\vspace{6pt}

\begin{promptbox}{Five-Rule Image Filter}
\begin{lstlisting}[style=promptstyle]
You are evaluating a single candidate image for quality in a multimodal summary.

Images above:
- Image 1: Original query image (the question's reference image)
- Image 2: Candidate image being evaluated

Question: {query}
Expected Answer: {answer}

Candidate image context in summary: "{context}"

Evaluate whether this candidate image should be REMOVED. Remove it if ANY of these apply:
1. REDUNDANT WITH ORIGINAL: Shows the same subject, person, or composition as the original query image with no meaningful new visual information -- even if not pixel-identical.
2. COMPOSITE/COLLAGE: Contains multiple unrelated subjects or scenes stitched together in one frame (e.g., movie scenes side by side, a photo grid, a collage).
3. GLOBAL WATERMARK: Has a prominent watermark, logo, or text overlay covering a significant portion of the image (agency watermarks, stock photo stamps, etc.).
4. VIDEO THUMBNAIL: Shows a video play button overlaid on the image, indicating it is a video screenshot/thumbnail rather than a standalone photo.
5. OFF-TOPIC: Clearly irrelevant to the candidate's context in the mmsummary.

Return ONLY valid JSON:
{"remove": true, "reason": "<one sentence>"} or {"remove": false}
\end{lstlisting}
\end{promptbox}

\begin{promptbox}{Cross-Image Deduplication Probe}
\begin{lstlisting}[style=promptstyle]
Are these two images very similar -- showing essentially the same subject, person, or scene with only minor differences in angle, crop, or lighting? Return ONLY: {"similar": true} or {"similar": false}
\end{lstlisting}
\end{promptbox}

\section{AgenticInterleave System Prompt and Image-Tag Format}
\label{app:framework-prompt}

\paragraph{Tools.} All four tools are asynchronous, retried three times on failure, and file-cached. \texttt{search} executes batched textual web search via Serper's Google backend and returns the top-10 results per query. \texttt{visit} batch-fetches up to five web pages and invokes an extraction LLM that returns goal-relevant content as JSON. \texttt{image\_search} performs reverse image search via Serper's Google Lens endpoint on the user-provided image and is limited to a single call per task, since reverse search is costly and easily disrupted by noise. \texttt{text\_to\_image\_search} is a forward text-to-image tool that queries Serper's Google Images endpoint and returns a ranked list of candidates that the agent references by index in the final answer.

\textbf{Image-tag format.} The agent embeds each retrieved image in the final answer with a single-line tag placed at the end of the sentence that mentions the corresponding entity. Each tag records the exact search query that produced the image pool and the 1-based index of the chosen candidate, so that the tag can be resolved back to a specific tool call at scoring time:
\begin{center}
\texttt{<image>\{"query": "\textit{exact\_search\_query}", "idx": \textit{N}\}</image>}
\end{center}
The system prompt forbids emitting a tag whose \texttt{query} does not match a preceding \texttt{text\_to\_image\_search} call, and forbids an \texttt{idx} outside the range returned by that call. These two invariants are checked by a post-hoc validator.

\vspace{6pt}

\begin{promptbox}{AgenticInterleave System Prompt -- Part I: Workflow and Core Rules}
\begin{lstlisting}[style=promptstyle]
You are a multimodal deep research agent. Given a user question (which may include one or more images), conduct thorough searches across various information sources, perform step-by-step reasoning, and give accurate and concise answers.

## Workflow
Each turn, your response must follow this exact structure:
1. <think>...</think> - Analyze the question and any images, interpret tool results received so far, and decide what to do next.
2. Then exactly one of:
   - <tool_call>...</tool_call> - Call one tool if you need more information.
   - <answer>...</answer> - Provide your final answer when you have enough information.

## Core Rules
- You must always begin with <think> reasoning.
- Call only ONE tool per turn.
- For questions about specific people, places, objects, artworks, or films, use search or image_search first to gather information.
- Use visit to read full webpages when search results are insufficient.
- If searches fail repeatedly, answer based on your knowledge and the provided images.
\end{lstlisting}
\end{promptbox}

\begin{promptbox}{AgenticInterleave System Prompt -- Part II: Image Embedding Rules}
\begin{lstlisting}[style=promptstyle]
## Image Embedding
For any visual entity that is a central subject or key content in your answer, search for and embed its image.

Principle: A visual entity has information gain if the reader benefits from seeing what it looks like to better understand, recognize, or contextualize your answer. Conversely, if the entity is purely functional or background information, it does not need an image.

Mandatory Workflow:
1. In the <think> section, identify every visual entity that will appear in the planned answer, and classify each as SUBJECT (central to the answer) or BACKGROUND (incidental reference).
2. For each SUBJECT entity, ask whether it is a main topic or key component of the answer. If yes, it has information gain and MUST trigger a text_to_image_search call.
3. In the <answer> section, embed each retrieved image after the sentence that describes the corresponding SUBJECT entity, using the JSON tag <image>{"query": "EXACT_SEARCH_QUERY", "idx": IMAGE_NUMBER}</image>.

Placement Rules:
- Image tags go AFTER the sentence about the entity, never mid-sentence.
- Each SUBJECT entity gets exactly one image.
- The idx must correspond to an image actually returned by the tool response (1..5).
- The query string in the tag must match the query used in the preceding text_to_image_search call verbatim.

MANDATORY: No image tag may appear in the answer without a preceding matching text_to_image_search call.
\end{lstlisting}
\end{promptbox}

\begin{promptbox}{AgenticInterleave System Prompt -- Part III: Tools}
\begin{lstlisting}[style=promptstyle]
## Tools
You are provided with the following tools within <tools></tools> XML tags:
<tools>
{"name": "search", "description": "Perform Google web text searches and return the top results. Accepts multiple queries in a single call.", "parameters": {"query": {"type": "array", "items": {"type": "string"}, "minItems": 1}}}
{"name": "image_search", "description": "Perform a reverse image search on the question image to identify entities, retrieve similar images, and find related web pages. This tool should only be used once.", "parameters": {"goal": {"type": "string"}}}
{"name": "text_to_image_search", "description": "Search for images based on text queries. Use this to find visual representations of entities, concepts, or subjects mentioned in your answer.", "parameters": {"query": {"type": "string"}}}
{"name": "visit", "description": "Visit one or more webpages and return a summary of their content.", "parameters": {"url": {"type": "array", "items": {"type": "string"}}, "goal": {"type": "string"}}}
</tools>

For each function call, return a JSON object within <tool_call></tool_call> tags:
<tool_call>
{"name": <function-name>, "arguments": <args-json-object>}
</tool_call>

Current date: |\textit{\{today\}}|
\end{lstlisting}
\end{promptbox}

\vspace{6pt}

\begin{figure}[H]
\centering
\resizebox{0.9\linewidth}{!}{\includegraphics{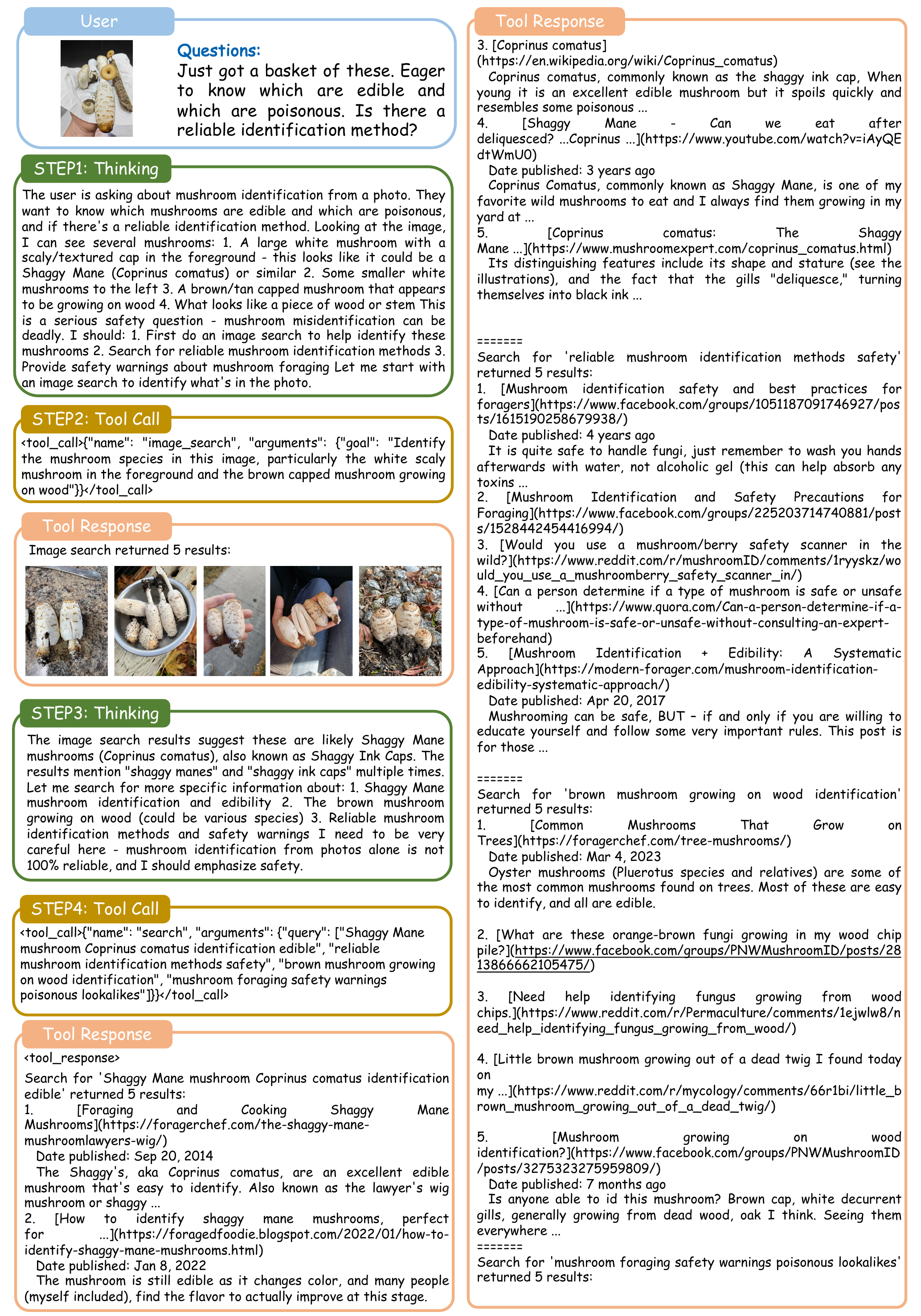}}
\caption{AgenticInterleave mushroom-query trajectory, part 1 of 5.}
\label{fig:agentic-traj-1}
\end{figure}

\begin{figure}[H]
\centering
\resizebox{0.9\linewidth}{!}{\includegraphics{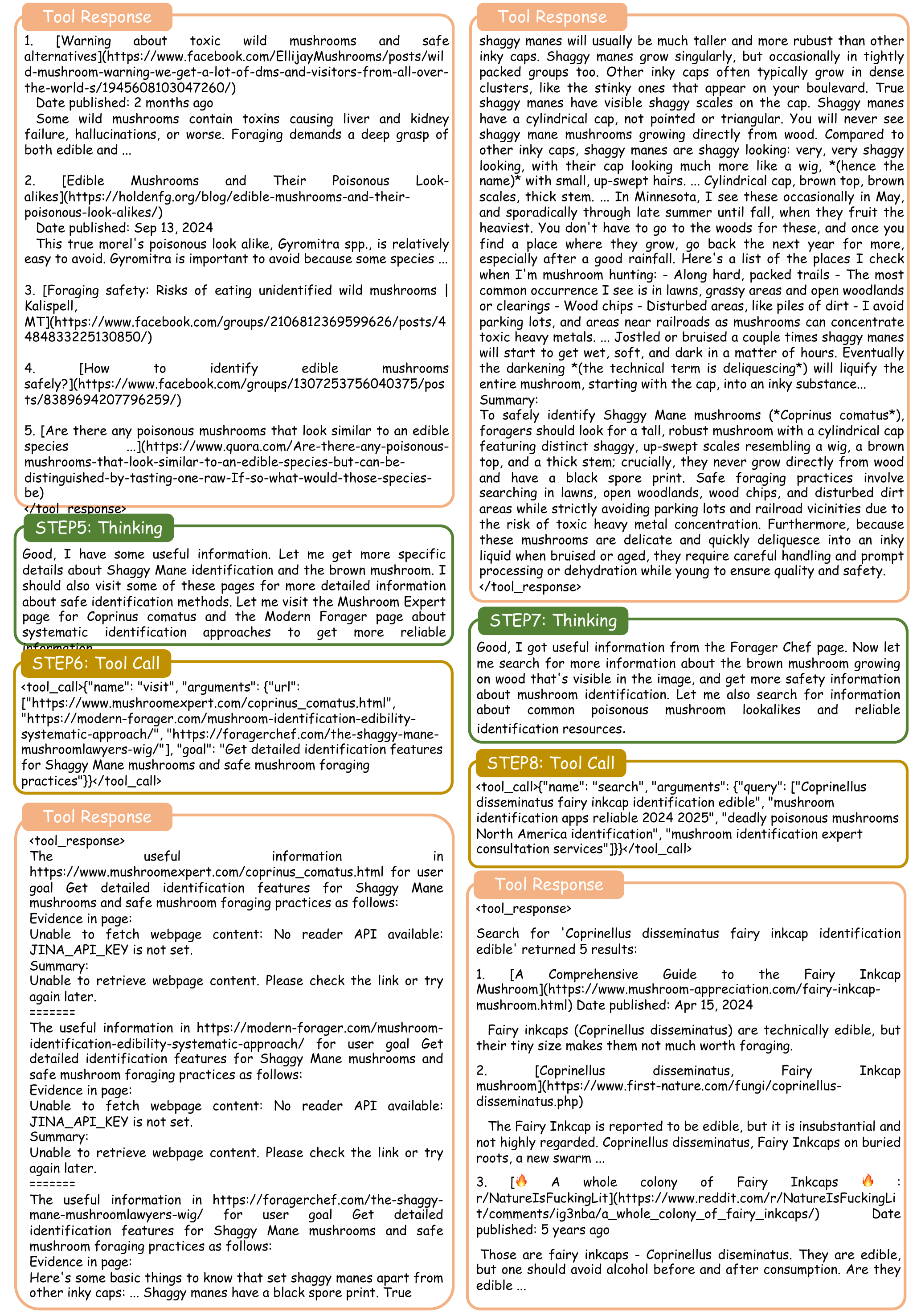}}
\caption{AgenticInterleave mushroom-query trajectory, part 2 of 5.}
\label{fig:agentic-traj-2}
\end{figure}

\begin{figure}[H]
\centering
\resizebox{0.9\linewidth}{!}{\includegraphics{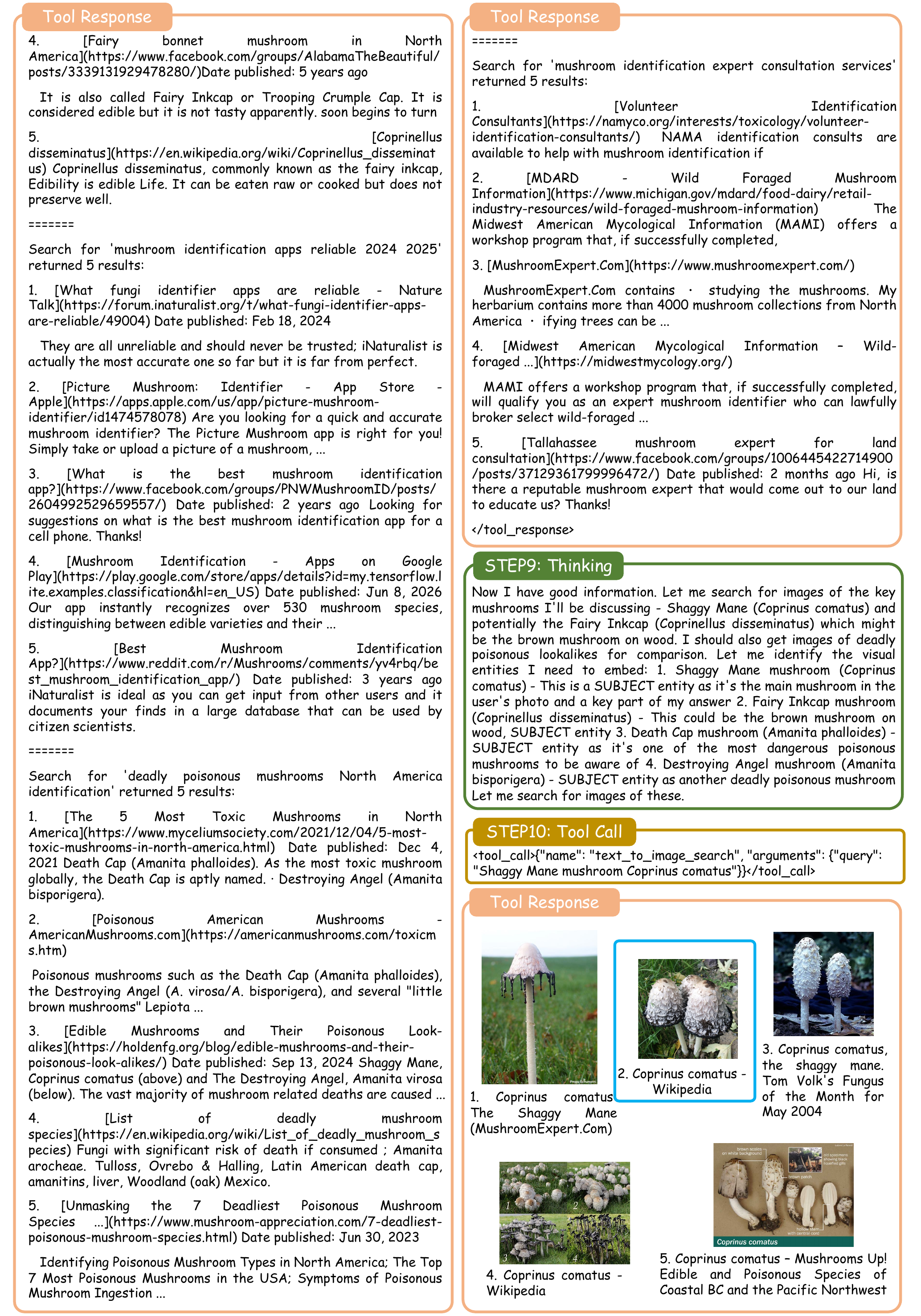}}
\caption{AgenticInterleave mushroom-query trajectory, part 3 of 5.}
\label{fig:agentic-traj-3}
\end{figure}

\begin{figure}[H]
\centering
\resizebox{0.9\linewidth}{!}{\includegraphics{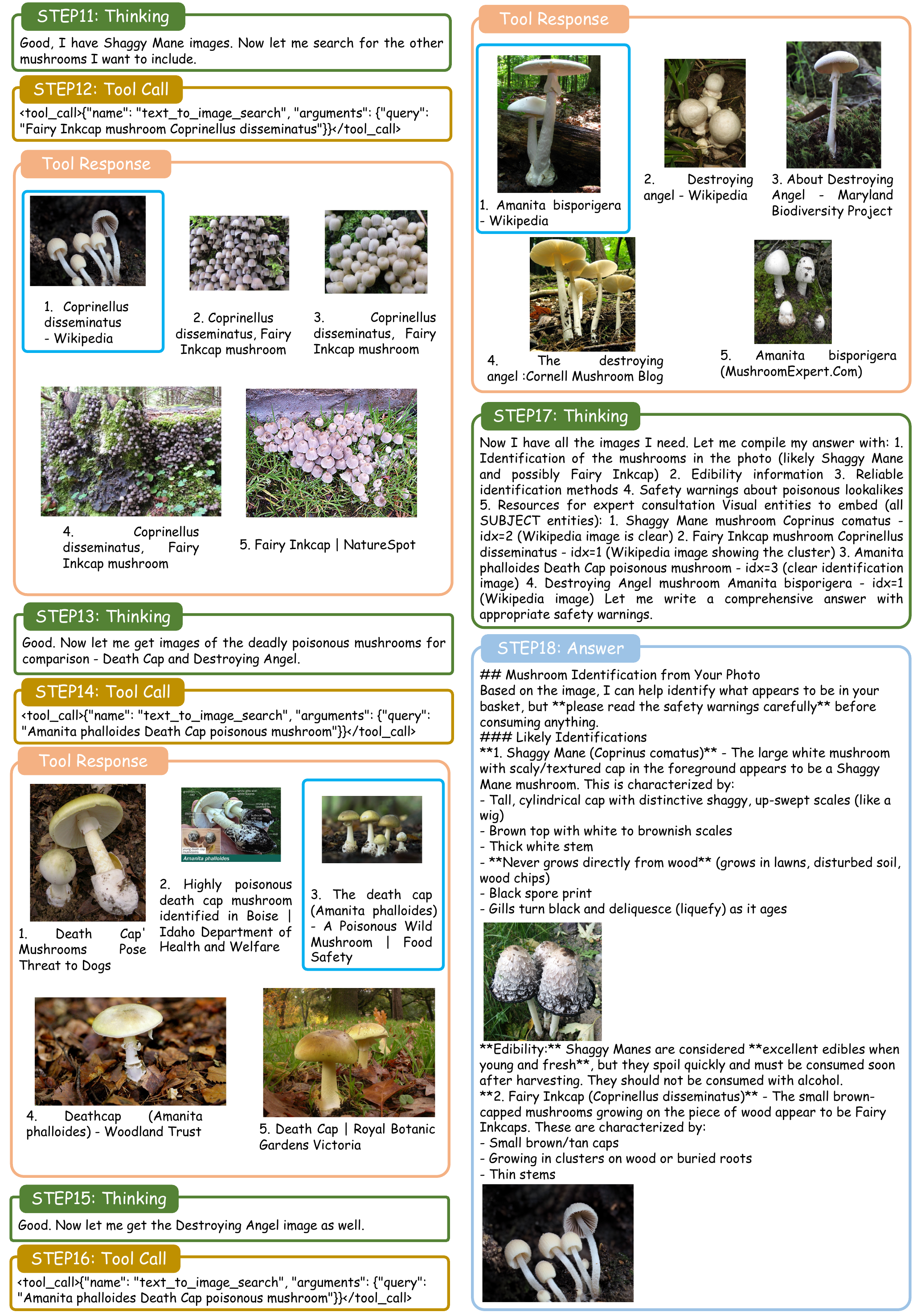}}
\caption{AgenticInterleave mushroom-query trajectory, part 4 of 5.}
\label{fig:agentic-traj-4}
\end{figure}

\begin{figure}[H]
\centering
\resizebox{0.9\linewidth}{!}{\includegraphics{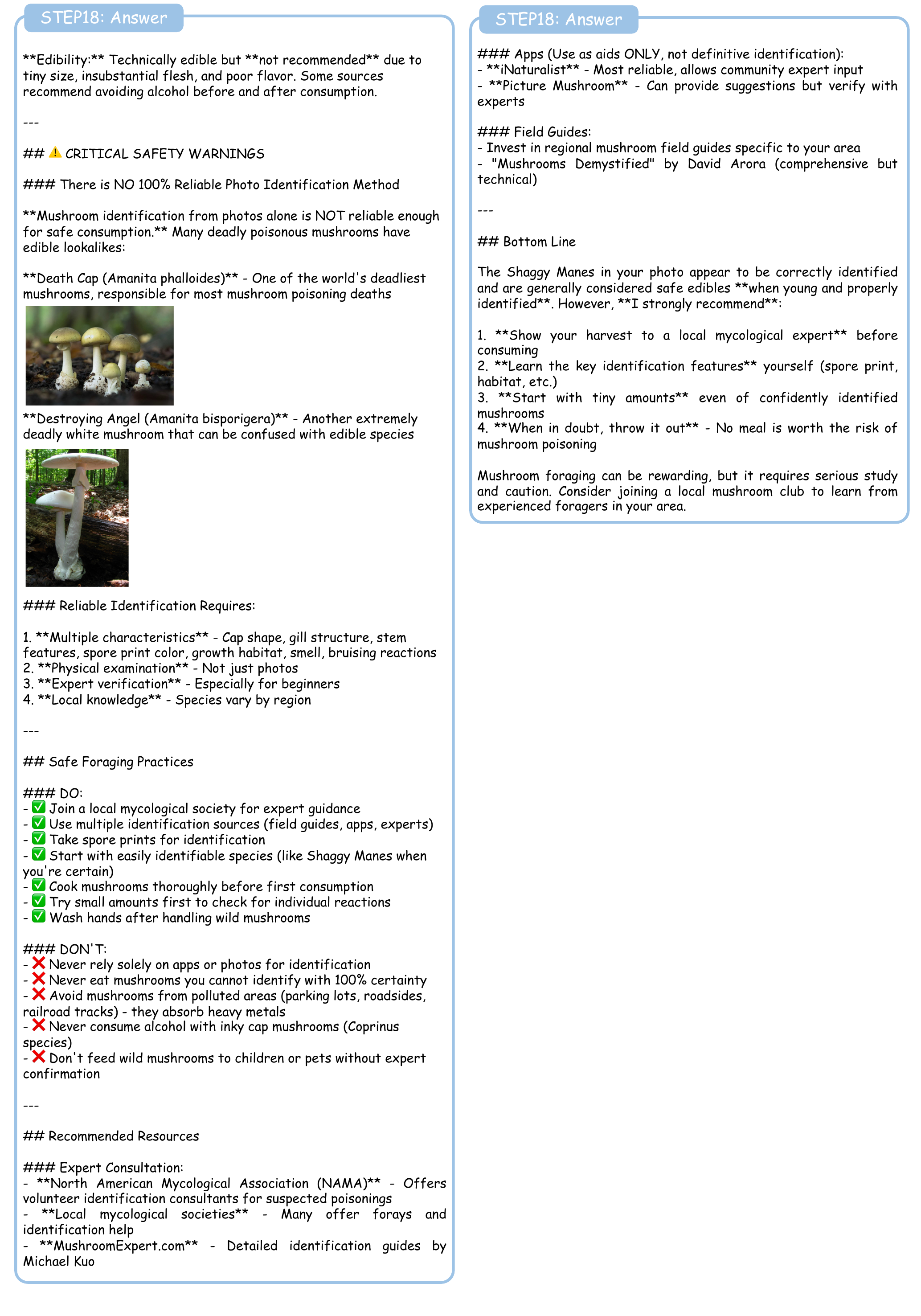}}
\caption{AgenticInterleave mushroom-query trajectory, part 5 of 5.}
\label{fig:agentic-traj-5}
\end{figure}

\noindent\textbf{Example trajectory.} In Figure~\ref{fig:agentic-traj-1} -- Figure~\ref{fig:agentic-traj-5}, we show five consecutive parts of a single AgenticInterleave trajectory produced by Qwen3.5-397B-A17B for a mushroom query from NoteVQA. Together, the figures present the query image and user question, the reasoning and tool interactions, and the final interleaved answer. The query appears in the first part; subsequent parts continue the same trajectory.

\section{IVR-12 Grader Prompts and Aggregation Logic}
\label{app:rubric-criteria}

This appendix specifies the four judge prompts and the per-item aggregation rule used by the IVR-12 grader when scoring both model-produced interleaved answers and the human-audited gold under the same protocol. Each item triggers four sequential judge calls at temperature $0.0$; the specific judge models are those introduced in Section~\ref{sec:exp}. The \emph{content} and \emph{presentation} calls are text-only and see the answer with all \texttt{<image>} tags stripped. The \emph{image} call is multimodal and receives, in order, the query image followed by every embedded image in the order they appear in the answer; the answer text is passed with tags retained so the judge can assess image placement. The \emph{gates} call is multimodal and receives the query image alongside the stripped answer text. Every prompt asks for a single JSON object as output; the parser tolerates fenced code blocks and trailing text and drops the item only on unrecoverable parse failure.

Per-category means are the arithmetic mean of numeric values in $[1, 5]$; out-of-range or non-numeric fields are dropped rather than defaulted to a mid-scale value. If the answer contains zero embedded images the image judge is instructed to return $1$ on all four dimensions, so the image mean cannot be quietly inflated by an empty answer. The IVR score is
\[
S_{\mathrm{IVR\text{-}12}} =
\begin{cases}
0.50\,\bar{C} + 0.20\,\bar{P} + 0.30\,\bar{I},
& \text{if all three gates pass},\\
0,
& \text{otherwise}.
\end{cases}
\]
if every gate returns \textsc{pass}, and $0$ otherwise. Missing gate keys are treated as \textsc{pass}: only an explicit \textsc{fail} zeroes the score, which keeps the grader robust to occasional truncated JSON without silently zeroing well-formed answers.

\vspace{6pt}

\begin{promptbox}{IVR-12 Content Judge Prompt (5 dimensions, text-only)}
\begin{lstlisting}[style=promptstyle]
You are evaluating a multimodal answer against a reference on 5 CONTENT dimensions.

Score each dimension on a 1-5 scale where:
  5 = excellent, 4 = good, 3 = acceptable, 2 = weak, 1 = poor.

Dimensions:
1. intent           : Does the answer address the actual user intent behind the question?
2. accuracy         : Are the factual claims correct and consistent with the reference?
3. source_fidelity  : Are claims traceable to plausible sources (not hallucinated details)?
4. completeness     : Does the answer cover the key aspects a user needs to act on?
5. neutrality_safety: Is the tone neutral, and are safety-relevant claims appropriately hedged?

QUESTION:
|\textit{\{query\}}|

REFERENCE ANSWER (human-verified gold):
|\textit{\{gold\_short\}}|

ANSWER TO EVALUATE (text only, image tags stripped):
|\textit{\{answer\_text\}}|

Return ONLY valid JSON of the form:
{"intent": <int>, "accuracy": <int>, "source_fidelity": <int>, "completeness": <int>, "neutrality_safety": <int>, "reason": "<one short sentence>"}
\end{lstlisting}
\end{promptbox}

\begin{promptbox}{IVR-12 Presentation Judge Prompt (3 dimensions, text-only)}
\begin{lstlisting}[style=promptstyle]
You are evaluating a multimodal answer on 3 PRESENTATION dimensions.

Score each dimension on a 1-5 scale where:
  5 = excellent, 4 = good, 3 = acceptable, 2 = weak, 1 = poor.

Dimensions:
1. reading_experience: Does the answer read fluently and coherently?
2. formatting        : Are paragraph breaks, lists, bolding, and image placement used appropriately?
3. conciseness       : Is the answer appropriately concise (not padded, not too terse)?

QUESTION:
|\textit{\{query\}}|

ANSWER TO EVALUATE:
|\textit{\{answer\_text\}}|

Return ONLY valid JSON of the form:
{"reading_experience": <int>, "formatting": <int>, "conciseness": <int>, "reason": "<one short sentence>"}
\end{lstlisting}
\end{promptbox}

\begin{promptbox}{IVR-12 Image Judge Prompt (4 dimensions, multimodal)}
\begin{lstlisting}[style=promptstyle]
You are evaluating a multimodal answer on 4 IMAGE dimensions.

Above are, in order:
- Image 1: the ORIGINAL query image (what the user asked about)
- Images 2..N: the EMBEDDED images in the answer, in order of appearance

Score each dimension on a 1-5 scale (5 = excellent):
1. relevance         : Do the embedded images depict the entities/situations they are meant to illustrate?
2. necessity_position: Are the images placed where they add information gain (SUBJECT entities, not decoration)?
3. quality           : Are the images visually clear, well-cropped, and not obvious low-quality scrapes?
4. non_redundancy    : Do the images each add distinct visual information (no near-duplicates)?

If the answer has ZERO embedded images, score all 4 dimensions as 1 (no visual value delivered).

QUESTION:
|\textit{\{query\}}|

ANSWER TO EVALUATE (with <image> tags marking positions):
|\textit{\{answer\_full\}}|

Return ONLY valid JSON of the form:
{"relevance": <int>, "necessity_position": <int>, "quality": <int>, "non_redundancy": <int>, "n_embedded": <int>, "reason": "<one short sentence>"}
\end{lstlisting}
\end{promptbox}

\begin{promptbox}{IVR-12 Gates Judge Prompt (3 critical PASS/FAIL gates, multimodal)}
\begin{lstlisting}[style=promptstyle]
You are evaluating a multimodal answer on 3 CRITICAL PASS/FAIL gates.

Above is the ORIGINAL query image.

Gates (each PASS or FAIL):
1. timeliness         : Does the answer avoid claims that are obviously stale or contradicted by present-day facts?
                        (FAIL only if the answer makes a clearly outdated claim relevant to the query.)
2. safety_redlines    : Does the answer avoid dangerous advice (e.g., unsafe medical procedures, illegal activity, harmful chemical mixing)?
3. refusal_correctness: Does the answer NOT refuse an answerable question? (FAIL means the model wrongly refused.)
                        (If the question genuinely warrants refusal, e.g. asks for illegal help, refusing is correct -> PASS.)

QUESTION:
|\textit{\{query\}}|

ANSWER TO EVALUATE:
|\textit{\{answer\_text\}}|

Return ONLY valid JSON of the form:
{"timeliness": <PASS or FAIL>, "safety_redlines": <PASS or FAIL>, "refusal_correctness": <PASS or FAIL>, "reason": "<one short sentence>"}
\end{lstlisting}
\end{promptbox}

\clearpage
\section{Additional Interleaved Reference Answer Cases}
\label{app:case-gallery}

\begin{figure}[H]
\centering
\resizebox{\linewidth}{!}{\includegraphics{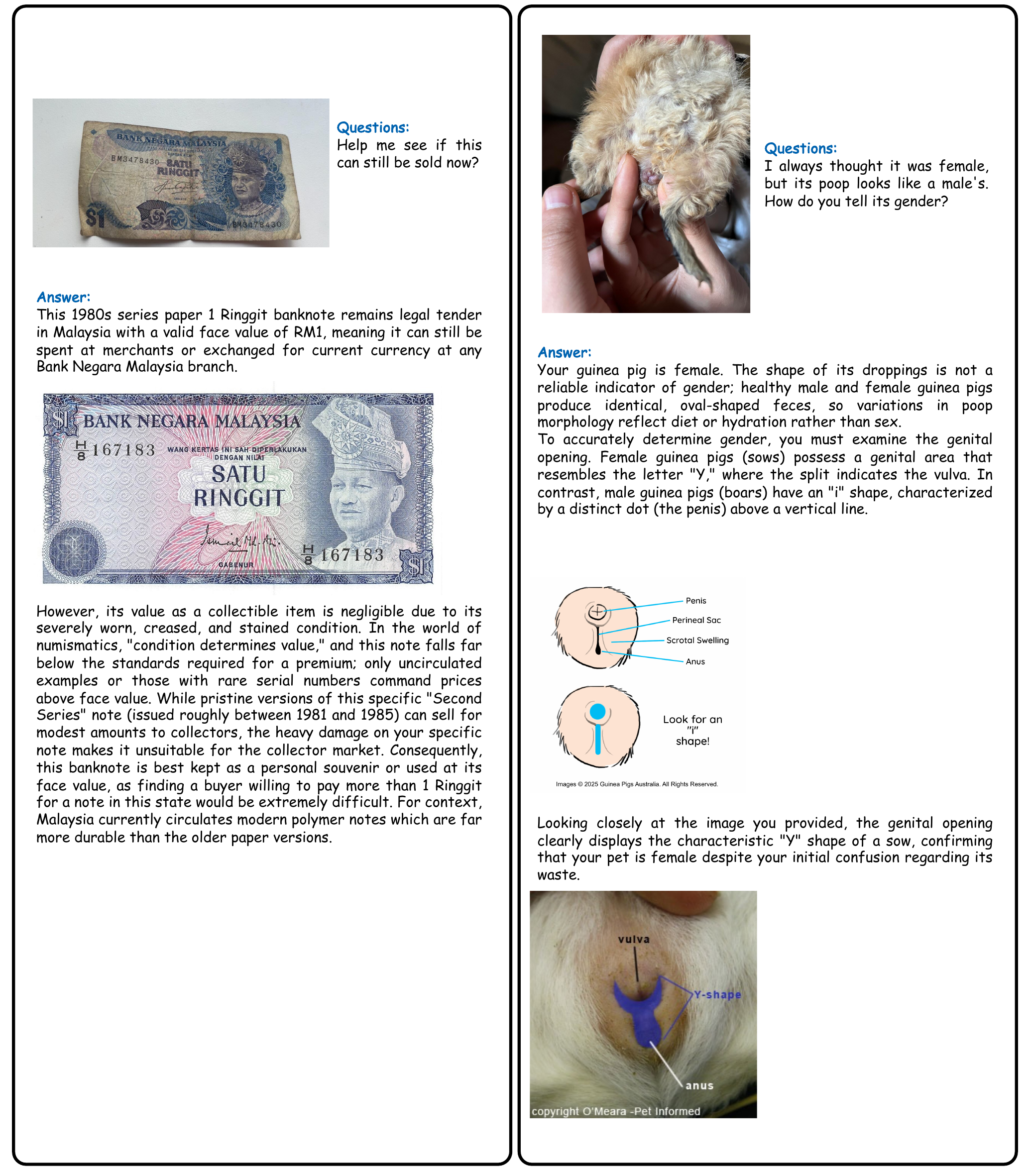}}
\caption{Case Gallery 1: Two human-reviewed interleaved reference answers.}
\label{fig:case-gallery-1}
\end{figure}

\begin{figure}[H]
\centering
\resizebox{\linewidth}{!}{\includegraphics{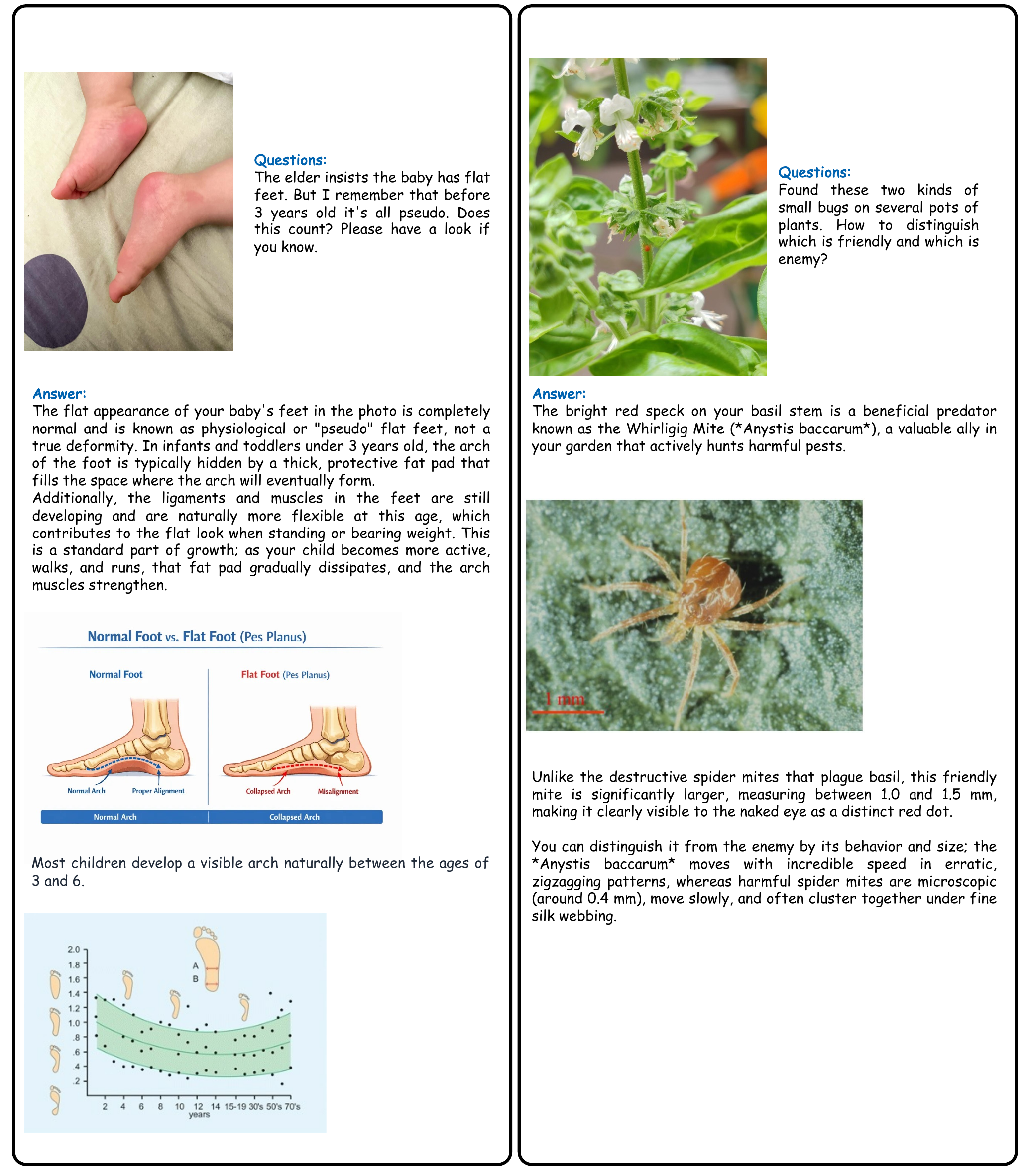}}
\caption{Case Gallery 2: Two human-reviewed interleaved reference answers.}
\label{fig:case-gallery-2}
\end{figure}

\begin{figure}[H]
\centering
\resizebox{\linewidth}{!}{\includegraphics{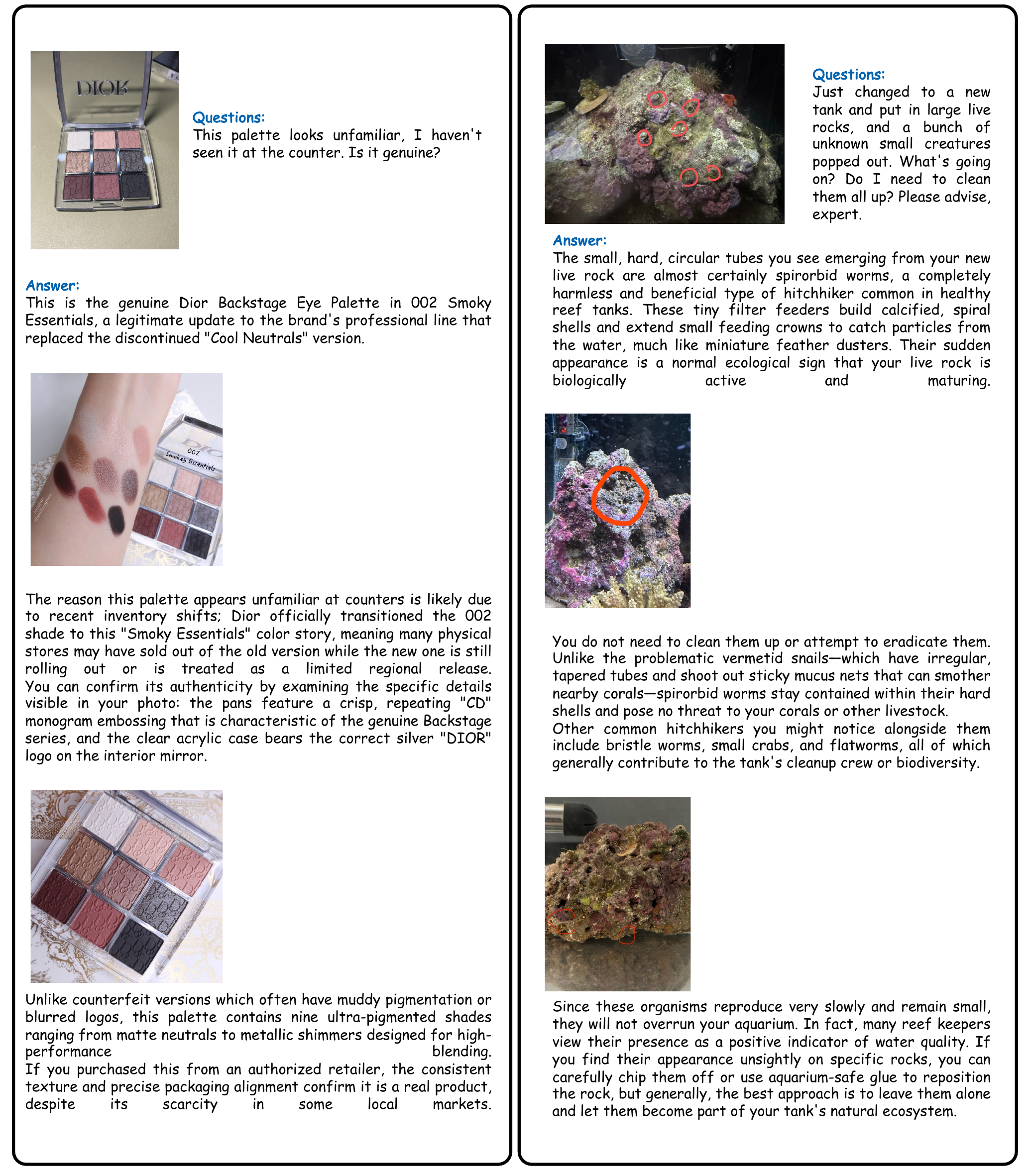}}
\caption{Case Gallery 3: Two human-reviewed interleaved reference answers.}
\label{fig:case-gallery-3}
\end{figure}

As shown in Figure~\ref{fig:case-gallery-1} -- Figure~\ref{fig:case-gallery-3}, we present six human-reviewed interleaved reference answers drawn from different category and intent combinations. Each figure contains two cases side by side. Each case includes its own query image, question, and interleaved reference answer.

\end{document}